\documentclass{article}  % For LaTeX2e
\usepackage{iclr2026_conference,times}

\usepackage{amsmath,amsfonts,bm}

\def\eqref#1{equation~\ref{#1}}
\def\1{\bm{1}}

\DeclareMathAlphabet{\mathsfit}{\encodingdefault}{\sfdefault}{m}{sl}
\SetMathAlphabet{\mathsfit}{bold}{\encodingdefault}{\sfdefault}{bx}{n}

\usepackage{capt-of, eucal}
\usepackage{xspace}
\usepackage{xcolor}
\usepackage{enumitem}
\usepackage{amsmath,amsthm,amssymb,amsfonts,dsfont,pifont,bm,bbm,mathrsfs,mathtools,nicefrac,extarrows,relsize}
\usepackage{algorithm,algpseudocode,listings}
\usepackage{booktabs,multirow,adjustbox,diagbox,threeparttable,tabularray,setspace}
\usepackage{wrapfig}

\definecolor{citeblue}{rgb}{0.21,0.49,0.74}
\usepackage[pagebackref=false,breaklinks,colorlinks,citecolor=citeblue,bookmarks=false]{hyperref}
\usepackage{subfigure}
\usepackage{wrapfig}
\usepackage[capitalize]{cleveref}  % Should be loaded after 'hyperref', and works perfectly with 'subfigure'.

\crefname{section}{Sec.}{Secs.}
\Crefname{section}{Section}{Sections}
\crefname{appendix}{Appendix}{Appendices}
\Crefname{appendix}{Appendix}{Appendices}
\crefname{table}{Table}{Tables}
\Crefname{table}{Table}{Tables}
\crefname{figure}{Fig.}{Figs.}
\Crefname{figure}{Figure}{Figures}
\crefname{equation}{Eq.}{Eqs.}
\Crefname{equation}{Equation}{Equations}
\crefname{theorem}{Thm.}{Thms.}
\Crefname{theorem}{Theorem}{Theorems}
\crefname{lemma}{Lem.}{Lems.}
\Crefname{lemma}{Lemma}{Lemmas}
\crefname{remark}{Rem.}{Rems.}
\Crefname{remark}{Remark}{Remarks}
\crefname{corollary}{Cor.}{Cors.}
\Crefname{corollary}{Corollary}{Corollaries}
\crefname{algorithm}{Alg.}{Algs.}
\Crefname{algorithm}{Algorithm}{Algorithms}
\definecolor{cellred}{RGB}{213, 123, 101}
\definecolor{cellgreen}{RGB}{0, 205, 0}
\definecolor{cellblue}{RGB}{54, 125, 189}
\definecolor{codegreen}{rgb}{0,0.6,0}
\definecolor{codegray}{rgb}{0.5,0.5,0.5}
\definecolor{codepurple}{rgb}{0.58,0,0.82}
\definecolor{backcolour}{rgb}{1.0,1.0,1.0}
\lstdefinestyle{mystyle}{
    backgroundcolor=\color{backcolour},
    commentstyle=\color{codegreen},
    keywordstyle=\color{magenta},
    numberstyle=\tiny\color{codegray},
    stringstyle=\color{codepurple},
    basicstyle=\ttfamily\scriptsize,
    breakatwhitespace=false,
    breaklines=true,
    captionpos=b,
    keepspaces=true,
    numbers=left,
    numbersep=5pt,
    showspaces=false,
    showstringspaces=false,
    showtabs=false,
    tabsize=2
}
\usepackage[most,skins,theorems]{tcolorbox}
\tcbset{
  aibox/.style={
    width=\linewidth,
    top=8pt,
    bottom=4pt,
    colback=blue!6!white,
    colframe=black,
    colbacktitle=black,
    enhanced,
    center,
    attach boxed title to top left={yshift=-0.1in,xshift=0.15in},
    boxed title style={boxrule=0pt,colframe=white,},
  }
}
\newtcolorbox{AIbox}[2][]{aibox,title=#2,#1}
\usepackage{amsmath} % For \text, \binom, \mathbb, \begin{cases}
\usepackage{amssymb} % For \mathbb
\usepackage{multirow}

\usepackage{rotating}  % in your preamble

\definecolor{demphcolor}{gray}{.2}

\definecolor{demphcolorinline}{gray}{.3}

\definecolor{demphcolor1}{gray}{.6}
\newcommand{\demphs}[1]{\textcolor{demphcolor1}{#1}}

\newcommand{\ie}{\textit{i.e.}\xspace}
\newcommand{\eg}{\textit{e.g.}\xspace}

\newcommand{\tocite}[1]{{\color{red} [TO CITE]}}

\newcommand{\metricname}{{ever\ pass\ rate\ }}

\newcommand{\dllm}{{dLLMs}\xspace}

\newcommand{\methodnamevoting}{{Temporal Self-Consistency Voting}\xspace}

\newcommand{\methodnamerft}{{Temporal Consistency Reinforcement}\xspace}

\title{ 
\includegraphics[width=0.06\textwidth]{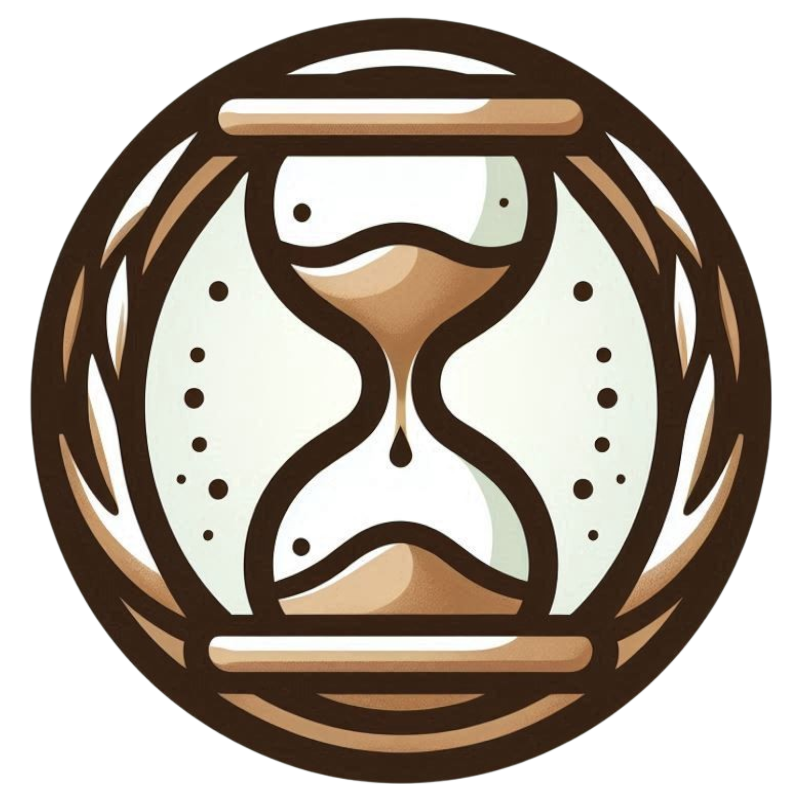}
Time Is a Feature: Exploiting Temporal Dynamics in Diffusion Language Models
}

\author{Wen Wang$^{1,2}$\thanks{WW and BF  contributed equally. CS is the corresponding author.},
~~
Bozhen Fang$^{1*}$,
~~
Chenchen Jing$^{1,3}$,
~~
Yongliang Shen$^{1}$,
~~
Yangyi Shen$^{4}$,
\\
\textbf{Qiuyu Wang}$^{2}$, 
~~
\textbf{Hao Ouyang}$^{2}$, 
~~
\textbf{Hao Chen}$^{1}$, 
~~
\textbf{Chunhua Shen}$^{1,3,2}$
\vspace{0.3cm}\\
$^1$ Zhejiang University~~
$^2$ Ant Group~~
$^3$ Zhejiang University of Technology~~
$^4$ Stanford University
}

\iclrfinalcopy % Uncomment for camera-ready version, but NOT for submission.

\begin{document}

\maketitle

\begin{abstract}

Diffusion large language models (dLLMs) generate text through iterative denoising, yet current decoding strategies discard rich intermediate predictions in favor of the final output. Our work here reveals a critical phenomenon, \textbf{temporal oscillation}, where correct answers often emerge in the middle process, but are overwritten in later denoising steps.
To address this issue, we introduce two complementary methods that exploit temporal consistency: 1) \textbf{\methodnamevoting}, a training-free, test-time decoding strategy that aggregates predictions across denoising steps to select the most consistent output; and 2) a post-training method termed \textbf{\methodnamerft}, which uses Temporal Semantic Entropy (TSE), a measure of semantic stability across intermediate predictions, as a reward signal to encourage stable generations. 
Empirical results across multiple benchmarks demonstrate the effectiveness of our approach.
Using the negative TSE reward alone, we observe a remarkable average improvement of \textbf{24.7\%} on the Countdown dataset over an existing dLLM. 
Combined with the accuracy reward, we achieve {absolute gains} of \textbf{2.0\%} on GSM8K, \textbf{4.3\%} on MATH500, \textbf{6.6\%} on SVAMP, and \textbf{25.3\%} on Countdown, respectively.
Our findings underscore the untapped potential of temporal dynamics in dLLMs and offer two simple yet effective tools to harness them.

Project webpage: \url{https://aim-uofa.github.io/dLLM-MidTruth}

\end{abstract}

\section{Introduction}

Diffusion large language models (\dllm)~\citep{nie2025large, zhu2025llada1.5, dream2025} have recently emerged as a 
promising alternative to the auto-regressive (AR) large language models, garnering significant attention for their competitive performance and potential for faster inference. 
In contrast to AR models, which generate text in a strictly sequential manner by predicting one token at a time, 
dLLMs operate through iterative cycles of denoising and remasking, predicting all masked tokens in parallel at each step.
A small subset of the predicted tokens, typically those with high confidence~\citep{nie2025large}, %is
are 
retained, while the remaining tokens are remasked and refined in subsequent steps.
\textit{Despite their drastic architectural differences, 
current \dllm typically adopt a decoding strategy that mirrors AR models:
solely relying on the sequence predicted in the final denoising step as the final answer, and discarding all the intermediate predictions. }

In this work, we challenge this convention by uncovering a critical phenomenon that we term \textbf{temporal oscillation}: correct answers often appear during intermediate denoising steps but are overwritten in later iterations. 
This discrepancy between the final output and intermediate correctness suggests that dLLMs possess rich temporal dynamics that are largely under-utilized.

\begin{figure*}
    \centering
    \includegraphics[width=\textwidth]{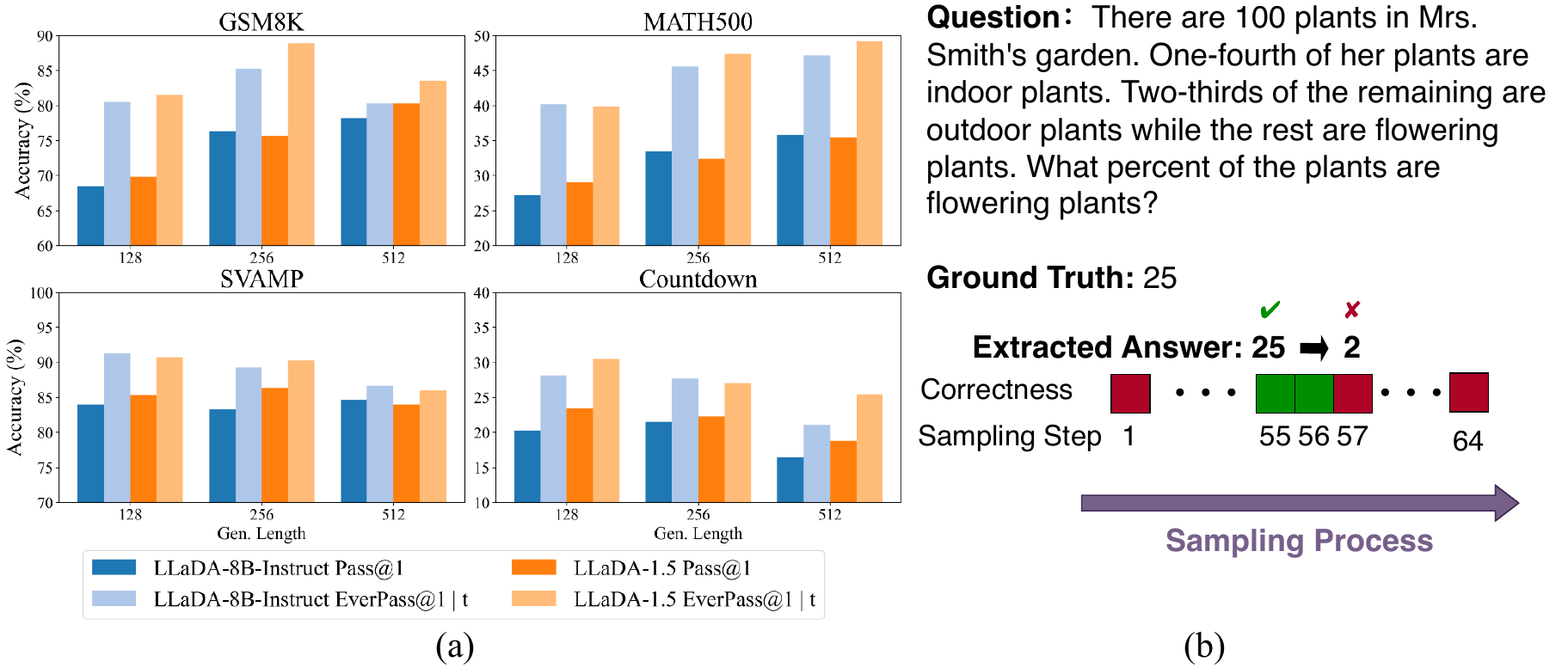}
    \caption{
        \textbf{Illustration of temporal oscillation during sampling.}
        (a) Across four datasets, a significant gap is observed between the \textbf{final answer's pass rate} (denoted as $\operatorname{Pass} @ 1$) and the \textbf{ever-pass rate at any intermediate step} (denoted as $\operatorname{EverPass} @ 1 \mid t$). This gap reveals the phenomenon we refer to as temporal oscillation, where correct intermediate answers are sometimes overwritten as the generation proceeds.
        (b) Example of temporal oscillation: For a given math problem, the model initially gives the correct answer, 25, at an intermediate step (\eg, step 55), aligning with the ground truth. 
        However, by the final step, this correct answer is replaced with an incorrect one: 2. 
        % More examples with detailed outputs can be found in \cref{app:subsec_examples_oscillation}.
    }
    \label{fig:temporal_oscillation}
\end{figure*}

As depicted in \cref{fig:temporal_oscillation}, we analyze two key metrics across four widely used benchmark datasets using two representative models: LLaDA-8B-Instruct~\citep{nie2025large} and LLaDA-1.5~\citep{zhu2025llada1.5}. The first metric, 
final pass rate,
measures the accuracy of the final output, while the second, ever-pass rate, captures whether a correct answer appears at any point during the decoding process.
In \cref{fig:temporal_oscillation}a, a consistent and significant discrepancy exists between these metrics. This gap reveals a critical phenomenon: models often generate correct answers during intermediate steps but subsequently overwrite them with incorrect ones.
\cref{fig:temporal_oscillation}b illustrates this concretely---in a math problem, the model produces the correct answer ``25" at sampling step 55, only to replace it with an incorrect ``2" by the final step 64.
More examples on temporal oscillation are presented in \cref{app:subsec_examples_oscillation}.

To better understand this behavior, we analyze \dllm from the entropy perspective and introduce a new metric: Temporal Semantic Entropy (TSE), which captures the distribution of semantic variation across intermediate outputs during decoding. 
Specifically, we collect the sequence of intermediate answers across denoising steps and group them into clusters based on semantic equivalence. 
TSE then quantifies the degree of uncertainty in the semantic content of these answers.
A higher TSE indicates greater semantic fluctuation throughout the trajectory, \ie, the model changes its answer frequently, while a lower TSE suggests convergence toward a stable meaning.

To harness the latent signals embedded in \dllm decoding, we treat temporal oscillation as an informative feature and develop two complementary methods that exploit the temporal dynamics:
\begin{itemize}
    \item \textbf{\methodnamevoting}: A training-free test-time decoding strategy that aggregates predictions across denoising steps and selects the most consistent output, considerably improving accuracy with negligible computational overhead.
    
    \item \textbf{\methodnamerft}: 
    A post-training method based on reinforcement learning that uses negative TSE as a reward signal to encourage stable and consistent generations.
    Importantly, leveraging negative TSE as the reward enables performance improvements without requiring ground-truth labels for reward computation.
\end{itemize}

Experiments across multiple datasets validate the effectiveness of both our decoding-time strategy and our RL-based post-training method. 
Specifically, \methodnamevoting brings an average improvement of \textbf{1.5\%}  over the LLaDA-8B-Instruct baseline with negligible overhead.
In terms of \methodnamerft, fine-tuning using the negative TSE reward alone, we observe a substantial average improvement of \textbf{24.7\%} on the Countdown dataset. 
When combined with the accuracy reward derived from ground truth, our approach yields notable improvements across diverse datasets: \textbf{2.0\%} on GSM8K, \textbf{4.3\%} on MATH500, \textbf{6.6\%} on SVAMP, and an impressive \textbf{25.3\%} on Countdown, respectively.
By quantifying and leveraging temporal consistency, we offer a new perspective on dLLM decoding and introduce practical tools to unlock their potential.
We hope that this study inspires further research into the temporal characteristics of diffusion decoding.

\section{Related Work}

\textbf{Diffusion language models.}
Building on the success of diffusion in image and video generation~\citep{song2020denoising, ho2020denoising, ho2022video}, diffusion methods have been extended to text. Early continuous approaches~\citep{han2022ssd, li2022diffusion} operate in continuous space, while others map text to the probability simplex~\citep{avdeyev2023dirichlet, stark2024dirichlet}. More recent work applies flow matching on the simplex to learn categorical distributions~\citep{davis2024fisher, cheng2024categorical}, but remains limited to simpler sequences.

Discrete diffusion models, pioneered by D3PM~\citep{austin2021structured}, advanced through masked token frameworks~\citep{shi2024simplified, sahoo2024simple, nie2024scaling} and scaling efforts. Lightweight variants like Plaid~\citep{gulrajani2023likelihood} and SEDD~\citep{lou2023discrete} rival GPT-2~\citep{radford2019language}, yet lag autoregressive models in scalability. To bridge this gap, BD3-LMs~\citep{arriola2025block} and Eso-LMs~\citep{sahoo2025esoteric} interpolate between autoregressive and diffusion paradigms, enabling parallel sampling with competitive performance. Recent efforts scale up \dllm: Dream~\citep{dream2025} converts pretrained autoregressive models into diffusion models, while LLaDA~\citep{nie2025large} trains strong models from scratch.

Another line of work studies sampling in \dllm. For instance, \citet{kim2025train} show token ordering affects performance and propose adaptive inference, while ReMDM~\citep{wang2025remasking} uses inference-time remasking to boost generation. Unlike previous work, we focus on improving \dllm from the underexplored perspective of temporal stability.

\textbf{Test-time Strategy.}
Test-time strategies~\citep{wei2022chain,madaan2023self,snell2024scaling,yao2023tree,liu2025rethinking} are widely used to improve LLM accuracy, consistency, and reliability. A simple yet effective method is Self-Consistency~\citep{wang2022self}, which selects the most consistent answer from multiple outputs via majority voting. Building on this idea, we propose a temporal self-consistency strategy tailored for \dllm, which adds negligible inference overhead and integrates seamlessly into existing frameworks.
Another important technique is semantic entropy~\citep{farquhar2024detecting,kuhn2023semantic}, an uncertainty metric that clusters semantically equivalent outputs before computing entropy. While previously applied to uncertainty estimation and hallucination detection, we extend it to \dllm by introducing Temporal Semantic Entropy, capturing stability and confidence throughout the denoising process.

\textbf{Post-Training using Reinforcement Learning.}
Group Relative Policy Optimization~\citep{shao2024deepseekmath,guo2025deepseek} (GRPO), a variant of Proximal Policy Optimization~\citep{schulman2017proximal}, computes advantages directly from group rewards, removing the need for a separately trained value function. GRPO has shown strong performance in reasoning tasks like mathematics and code generation, and promising results across broader modalities~\citep{huang2025vision,shen2025vlm,qi2025vln,zhong2025omni,li2025temporal,damani2025beyond}. Building on this, refinements such as DAPO~\citep{yu2025dapo} introduce dynamic sampling to balance training batches, while entropy-based methods~\citep{zhang2025right,prabhudesai2025rent,cui2025entropy,agarwal2025unreasonable} further enhance RL. For example, EMPO~\citep{zhang2025right} derives rewards from semantic entropy, and Seed-GRPO~\citep{chen2025seed} improves advantage estimation. Recent adaptations to \dllm include \textit{diffu}-GRPO~\citep{zhao2025d1}, UniGRPO~\citep{yang2025mmada}, and coupled-GRPO~\citep{gong2025diffucoder}, which still rely on ground-truth rewards. By contrast, our method is fully unsupervised, enhancing temporal consistency without ground-truth supervision.

\section{Explorations on \dllm}

\subsection{Preliminaries on \dllm} 
DLLMs formulate text generation as a process of iteratively denoising text sequences across different time steps. 
Let $ x_0 \sim p_{\text{data}}(x_0) $ denote the original input sequence, and $ x_t $ represent its noisy version at time $ t \in [0, T] $. Typically, the noise is introduced by masking a portion of tokens in the original training data.
The forward noising process is defined as a Markov chain $ q(x_{1:T} \mid x_0) = \prod_{t=1}^T q(x_t \mid x_{t-1}) $, which progressively adds noise to $ x_0 $ over time steps. 
This process incrementally transforms the clean sequence $ x_0 $ into a highly noisy version $ x_T $ through a series of conditional transitions.
In contrast, the reverse (generative) process is modeled as:
\begin{equation}
\label{eq:dllm}
\resizebox{0.7\columnwidth}{!}{$
p_\theta(x_{0:T}) = p_\theta(x_T) \prod_{t=1}^T p_\theta(x_{t-1} \mid x_t) = \prod_{t=1}^T q(x_{t-1} \mid x_0) p_\theta(x_0 \mid x_t).
$}
\end{equation}
This generative process can be divided into two key steps. 
The first step involves $ p_\theta(x_0 \mid x_t) $, a model trained to predict the clean text sequence based on the current noisy input. 
We denote the answer decoded at the $t$-th step as $x_0^t$, \ie, $ x_0^t = p_\theta(x_0 \mid x_t) $.
The second step relies on $ q(x_{t-1} \mid x_0) $, referred to as the re-masking process. 
This process applies the forward procedure to the currently predicted $x_0^t$, yielding a sequence $x_{t-1}$ that is less noisy compared to $x_t$. 
Various strategies can be employed for this process, such as random or low-confidence re-masking, as in \citep{nie2025large}.

\subsection{Temporal Oscillation}

In the reverse process of \dllm, predictions $p_\theta(x_0 \mid x_t)$ at a single step are often inaccurate, especially under high noise when $t$ is large. 
Existing models perform iterative denoising according to the generative process framework, where the final output is determined by the prediction $p_\theta(x_0 \mid x_1)$ at the last denoising step, while neglecting all intermediate predictions $\{x_0^t = p_\theta(x_0 \mid x_t)\}_{t=2}^{T}$ generated during the iterative process. In this work, we conduct an in-depth investigation into these intermediate step results, revealing a critical phenomenon in diffusion-based text generation.

To formalize our analysis, let $e_{i,k}$ denote the $\operatorname{Pass} @ 1$ rate~\citep{chen2021evaluating} for the prediction generated at the $k$-th noise step on the $i$-th question in the evaluation benchmark. 
Building on this, we introduce the \metricname, denoted as $\operatorname{EverPass} @ 1 \mid t$, which measures the proportion of questions in the dataset for which the model produces a correct answer at \textit{any} timestep along the sampling trajectory. Formally, it is defined as:
\begin{equation}
\resizebox{0.35\columnwidth}{!}{$
\operatorname{EverPass} @ 1 \mid t=\underset{i}{\mathbb{E}}
\left\{ \underset{k \in \{1, \dots, t\}}{\max} e_{i,k} \right\}
$}
\end{equation}
This metric captures the cumulative correctness across all sampling steps, reflecting the overall fraction of questions for which the model arrives at a correct solution at least once, even if that solution is later discarded in the final output.

\textbf{Experiment Setup.}
We compare the final pass rate, \ie, $\operatorname{Pass}@1$ at the last step with $\operatorname{EverPass}@1 \mid t$ on two representative \dllm: LLaDA-8B-Instruct and LLaDA-1.5, evaluated across different answer lengths and four reasoning benchmarks:
GSM8K~\citep{cobbe2021training}, MATH500~\citep{lightman2023let}, SVAMP~\citep{patel2021nlp}, and Countdown~\citep{tinyzero}.

\begin{figure*}
    \centering
    \includegraphics[width=\textwidth]{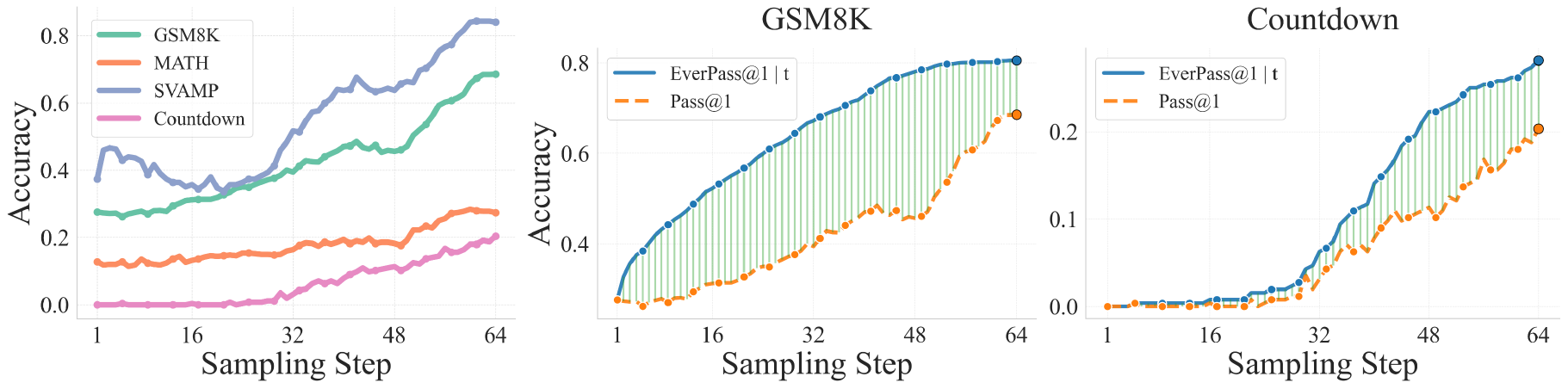}
    \caption{
        \textbf{Patterns of accuracy evolution over diffusion sampling steps.}  
        Responses of length 128 are generated with 64 steps using LLaDA-8B-Instruct.  
        \textbf{Left}: Accuracy generally rises with more steps across datasets; SVAMP starts high, while harder ones like Countdown start low but improve steadily. 
        \textbf{Middle/Right}: We compare the final pass rate, $\operatorname{Pass}@1$, with cumulative $\operatorname{EverPass}@1 \mid t$ over steps.  
        A clear gap persists between them, shown by the green shaded area.  
    }
    \label{fig:accuracy_analysis}
\end{figure*}

\textbf{Observations.}
As shown in \cref{fig:temporal_oscillation} and \cref{tab:main_voting}, there is a notable gap between the final pass rate and the \metricname. 
For instance, on GSM8K with length 128, LLaDA-8B-Instruct achieves 68.5\% final pass rate versus 80.5\% \metricname, a gap of 12.0\%. 
This gap shows that many questions are correctly solved at intermediate steps but later revised to incorrect answers during refinement. 
It reveals an instability in iterative decoding, where correct paths can be overwritten as generation proceeds. 
We term this phenomenon \textit{temporal oscillation}, with examples in \cref{app:subsec_examples_oscillation}.

\begin{AIbox}{Takeaway 1: Correct intermediate answers are lost during sampling}
During sampling, answers may oscillate between correct and incorrect states across diffusion steps. 
A notable portion of questions achieve correct answers in the intermediate steps, but ultimately yield incorrect results in the final step.
\end{AIbox}

\subsection{Analyses}
\label{subsec:observations_analy}

To gain a deeper understanding of the temporal oscillation phenomenon, we conduct comprehensive analyses from multiple dimensions: accuracy, entropy, and semantic stability across decoding steps.

\textbf{Accuracy Across Sampling Steps.}
As shown in \cref{fig:accuracy_analysis}a, accuracy generally improves with more sampling steps. Simpler datasets like SVAMP start high and remain stable, while harder ones like Countdown begin low but benefit from iterative refinement.
To probe further, we compare $\operatorname{Pass}@1$ and $\operatorname{EverPass}@1 \mid t$ on GSM8K and Countdown. Early correct predictions appear sooner on GSM8K but later on Countdown, and a growing gap between the two metrics reveals that early correctness does not ensure stable reasoning. This underscores the importance of preserving correct intermediate states. Additional results for SVAMP and MATH500 are in \cref{app:subsec_math_and_svamp}.

\begin{figure*}
    \centering
    \includegraphics[width=\textwidth]{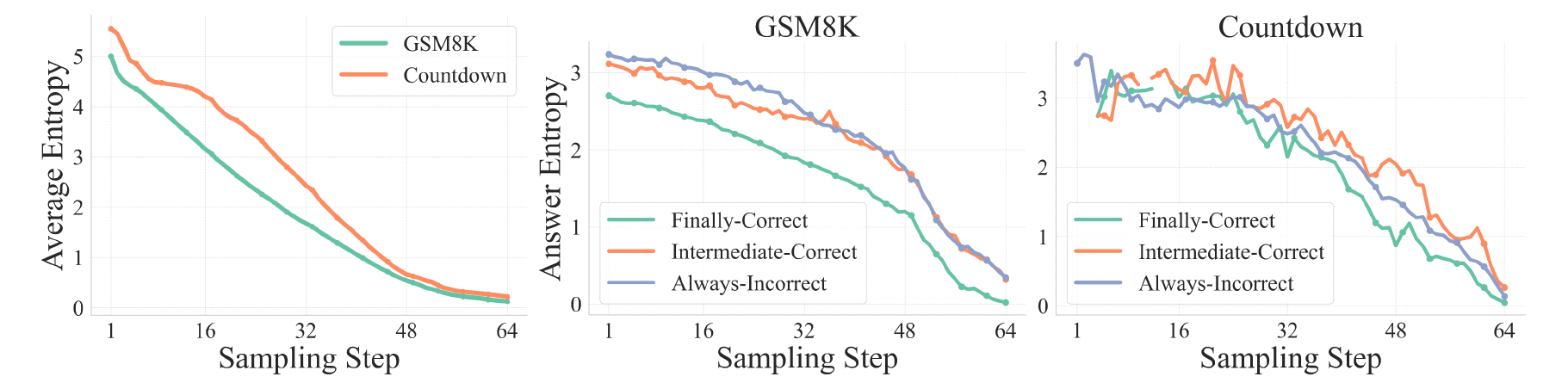}
    \caption{
        \textbf{Patterns of entropy evolution over diffusion sampling steps.} 
        Responses are generated with length 128 using 64 diffusion steps from the LLaDA-8B-Instruct model.
        \textbf{Left:} Average token-level entropy decreases steadily during sampling. GSM8K shows lower entropy than Countdown, aligning with its higher accuracy.
        \textbf{Middle and Right: } Both Intermediate-Correct and Always-Incorrect questions exhibit higher overall entropy compared to Finally-Correct ones. On GSM8K, Intermediate-Correct questions display lower entropy in the early steps than Always-Incorrect, indicating initial confidence, whereas on Countdown the entropy trend is less stable.
    }
    \label{fig:entropy_analysis}
\end{figure*}

\begin{wrapfigure}{r}{0.5\textwidth}
    \centering
    \includegraphics[width=0.43\textwidth]{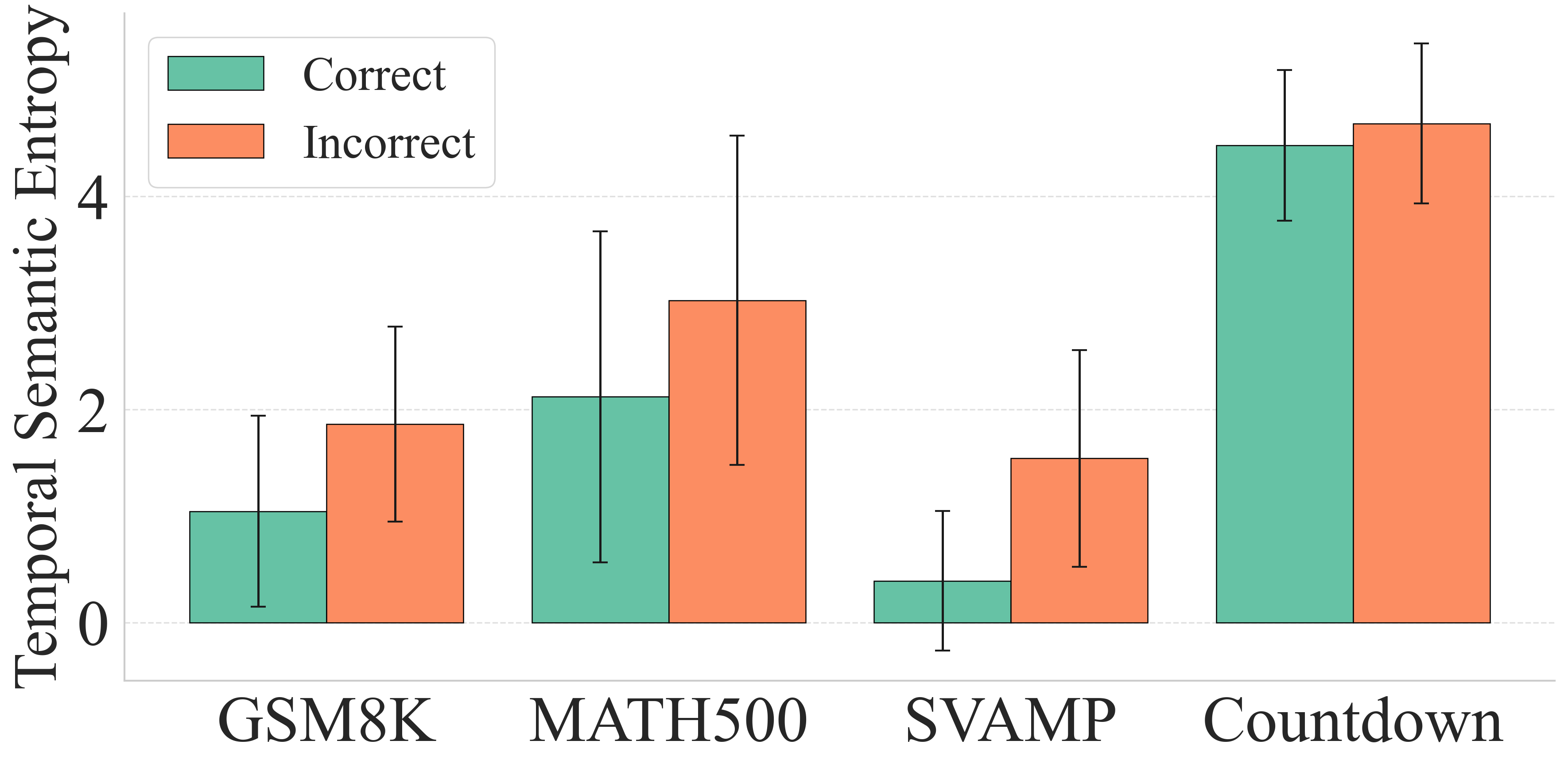}
    \caption{
    \textbf{Temporal semantic entropy across four benchmarks.}
    This metric measures the uncertainty in the semantic content of answers across decoding steps. Statistically, correctly answered questions exhibit lower entropy.
    }
    \label{fig:temporal_entropy}
\end{wrapfigure}

\textbf{Entropy Across Sampling Steps.}
Temporal oscillations reflect model uncertainty. To quantify this, we analyze the average token-level entropy. As shown in \cref{fig:entropy_analysis}, entropy decreases steadily and approaches zero by the final step, with GSM8K starting lower, indicating higher initial confidence.
We categorize questions into three groups: Finally-Correct, Always-Incorrect, and Intermediate-Correct, where Intermediate-Correct means at least one intermediate step is correct but the final answer is incorrect.
For GSM8K and Countdown, incorrect final answers show consistently higher entropy. On GSM8K, Intermediate-Correct questions begin with lower entropy than Always-Incorrect, suggesting initially confident but unstable predictions.
Considering the semi-autoregressive sampling strategy~\citep{nie2025large}, we also measure the average entropy of the currently generated block, detailed in \cref{app:subsec_block_entropy}.

\textbf{Temporal Semantic Entropy.}
Token-level entropy reflects local uncertainty, but we also need a measure of semantic consistency across the decoding trajectory.
We therefore introduce \textit{Temporal Semantic Entropy} (TSE), which captures the semantic variations of answers during sampling.

During decoding, we obtain a sequence of $T$ intermediate answers, denoted by $\{x_0^t\}_{t=1}^{T}$. We cluster them by semantic meaning into $\mathcal{C}=\{C_1,\ldots,C_K\}$, 
where each cluster $C_k$ groups answers with equivalent semantics.
% where each cluster $C_k = \{x_0^t : \text{meaning}(x_0^t) = k\}$ contains all answers with equivalent semantics (outcome equals $k$).
The TSE of a trajectory is defined as:
\begin{equation}
\label{eq:temporal_semantic_entropy}
\resizebox{0.6\columnwidth}{!}{$
\operatorname{TSE}(\{x_0^t\}_{t=1}^{T})
=-\sum_{C_k}\Big(\!\!\sum_{x_0^t \in C_k} p(x_0^t)\Big)\log\Big(\!\!\sum_{x_0^t \in C_k} p(x_0^t)\Big),
$}
\end{equation}
which quantifies the uncertainty in semantic content across steps: higher TSE indicates more semantic variation, while lower TSE implies convergence to a consistent meaning.

As shown in \cref{fig:temporal_entropy}, TSE offers insight into model behavior during generation.
In datasets like Countdown and MATH, where performance is weaker, we observe higher entropy than in GSM8K and SVAMP, reflecting greater semantic instability.
Moreover, questions ultimately answered correctly generally exhibit lower entropy than incorrect ones, indicating that stable, semantically consistent trajectories align with better performance.
Thus, high TSE may signal model uncertainty and highlight samples for further improvement. More results on TSE are provided in \cref{app:subsec_tse_vary_len}.

\begin{AIbox}{Takeaway 2: Correct answers statistically exhibit lower temporal semantic entropy}
Temporal semantic entropy, computed over intermediate predictions during the decoding trajectory, reflects the semantic stability of the model’s outputs. Statistically, correctly answered questions tend to have lower entropy, indicating greater consistency and confidence throughout the generation process.
\end{AIbox}

\section{Method}

\subsection{\methodnamevoting}
\label{subsec:method_voting}
We propose a temporal self-consistency decoding strategy for \dllm, which leverages intermediate predictions to improve final outputs.
As discussed in \cref{subsec:observations_analy}, while the last timestep usually yields the best result, the correct answer may also appear earlier, so relying only on the final prediction risks discarding better outputs.
To address this, we aggregate predictions across timesteps using a weighted voting mechanism.
Formally, given a diffusion sampling trajectory $\left\{x_0^t\right\}_{t=1}^{T}$, our method selects the final answer $a^*$ according to a weighted vote over all timesteps:
\begin{equation}
\resizebox{0.5\columnwidth}{!}{$
a^* = \arg \max _a \sum_{t=1}^T f(t) \cdot \mathbbm{1} \left(\text { meaning }\left(x_0^t\right) = a\right).
$}
\end{equation}
Here, $\mathbbm{1}(\cdot)$ indicates whether $x_0^t$ decodes to $a$, and $f(t)$ is a timestep weighting function.
Since accuracy generally increases with later steps, we design $f(t)$ as a monotonically decreasing function of diffusion step.
We experiment with constant, linear, and exponential weighting, as in \cref{subsec:exp_vote}.

\noindent\textbf{Discussion.}
Our method is conceptually related to self-consistency decoding \citep{wang2022self}, which improves reasoning in autoregressive LLMs by sampling diverse reasoning paths and selecting the most consistent answer via majority voting. 
However, self-consistency requires multiple full-length forward passes with high cost. 
% leading to significant computational overhead.
In contrast, our approach requires only a single sampling trajectory. 
By exploiting the temporal nature of diffusion inference indicated by \cref{eq:dllm}, we obtain a series of intermediate predictions without additional model evaluations. 
This makes our method both efficient and effective for boosting accuracy through temporal aggregation.

\subsection{\methodnamerft}
\label{subsec:method_rl}

Motivated by our observation in \cref{subsec:observations_analy} that correct answers generally exhibit lower Temporal Semantic Entropy (TSE) than incorrect ones, reflecting stronger semantic consistency over time, we propose a post-training approach designed to encourage temporal consistency in model outputs.
Specifically, we adopt Group Relative Policy Optimization (GRPO)~\citep{shao2024deepseekmath, guo2025deepseek} as our reinforcement learning framework and use TSE as a self-supervised reward signal.

\textbf{Negative TSE as the Reward.}
Following GRPO, for each question $q$ sampled from the dataset $\mathcal{D}$, we draw a group of $G$ responses $\{o_1, o_2, \ldots, o_G\}$ from the old policy $\pi_{\theta_{\text{old}}}$. 
Each response $o_i$ receives a scalar reward $r_i = -\operatorname{TSE}(o_i)$, 
where $\operatorname{TSE}(o_i)$ is computed using \cref{eq:temporal_semantic_entropy} from \cref{subsec:observations_analy}.
This reward encourages the model to produce responses whose intermediate predictions remain semantically consistent throughout the decoding process.
Based on this, we define the unnormalized advantage~\citep{liu2025understanding} for all tokens $k = 1, \ldots, |o_i|$ as
$
A_i^k(\pi) = r_i(\pi) - \operatorname{mean}\left(\left\{r_j(\pi)\right\}_{j=1}^G\right).
$
The training objective follows the standard GRPO formulation with our TSE-based reward:
\begin{equation}
\label{eq:grpo_loss}
\resizebox{0.93\columnwidth}{!}{
$
\displaystyle
\mathcal{L}_{\text{GRPO}}(\theta)
= \mathbb{E}_{\stackrel{q \sim \mathcal{D}}{o_1, \ldots, o_G \sim \pi_{\theta_{\text{old}}}(\cdot | q)}}
    \left[ \left( \frac{1}{G}\sum_{i=1}^G \frac{1}{|o_i|}\sum_{k=1}^{|o_i|} \min
    \left(\rho_i^k A_i^k, \text{clip}\left(\rho_i^k, 1-\varepsilon, 1 + \varepsilon \right) A_i^k\right) \right)
- \beta D_{\text{KL}}\left[\pi_\theta(\cdot | q)\| \pi_\text{ref}(\cdot | q) \right]
\right],
$
}
\end{equation}
where
$\pi_{\text{ref}}$ is the reference policy, $\rho_i^k = \frac{\pi_\theta(o_i^k \mid q)}{\pi_{\theta_{\text{old}}}(o_i^k \mid q)}$ is the importance sampling ratio, $\varepsilon$ is the clipping threshold, and $\beta$ controls the strength of the KL penalty.
To compute the token-level probabilities used in $\rho_i^k$, we follow the \textit{diffu}-GRPO method~\citep{zhao2025d1}, which estimates these probabilities by averaging outputs from multiple randomly masked versions of the prompt.

\textbf{Combining TSE with Accuracy Reward.}
When ground-truth answers are available during training, we combine TSE with an accuracy reward to further enhance performance. Specifically, We design a composite reward function that integrates correctness and temporal consistency. 

The accuracy reward follows a binary scheme: it assigns 1 when the model's prediction is correct ($o_i = o^*$), and 0 otherwise. 
To measure temporal consistency, we derive a normalized confidence score from TSE: $c(o_i) = \frac{\mathcal{H}_{\max} - \operatorname{TSE}(o_i)}{\mathcal{H}_{\max}}$, $\mathcal{H}_{\max} = \log T$.
Here, $T$ denotes the total sampling steps. 
This normalization yields $c \in [0,1]$, with larger values indicating stronger consistency.

We further transform $c(o_i)$ using the spherical scoring rule~\citep{gneiting2007strictly}, which has shown strong empirical performance across tasks.
The final reward is defined as:
\begin{equation}
\label{eq:reward_combine}
r_i = \mathbbm{1}_{o_i = o^*} + \frac{c(o_i)}{\sqrt{(c(o_i))^2  + (1 - c(o_i))^2}}.
\end{equation}
Here, the first term enforces correctness while the second term encourages higher temporal consistency. 
This formulation makes the reward sensitive not only to prediction accuracy, but also to the stability of the underlying generation process.
A detailed comparison of various scoring rules and their empirical performance in the \cref{app:subsec_scoring_rules}.

\noindent\textbf{Discussion.}
Unlike prior reinforcement learning post-training methods for \dllm, such as \textit{diffu}-GRPO, which relies on ground-truth answers for reward computation, our approach operates without any labeled data. 
Instead, we harness the model's internal temporal dynamics as a self-supervised signal, employing negative TSE to assess answer quality. 
This enables our method to be broadly applicable, particularly in unsupervised settings, and offers a novel direction for improving \dllm.
Furthermore, we show that combining the negative TSE reward with the accuracy reward based on ground-truth answers yields notably better performance than using the accuracy reward alone.
\section{Experiments}
\label{sec:exp}

\begin{table}[t] 
\centering
\caption{
\textbf{Performance of temporal majority voting.}
We compare three strategies: fixed, linear, and exponential weighting, on four datasets using LLaDA-8B-Instruct and LLaDA-1.5. 
Bold numbers mark group bests, and \textcolor{green!70!black}{green values} show gains over baseline. For reference, we report the oracle $\operatorname{EverPass}@1 \mid t$ as an upper bound.
}
\label{tab:main_voting} 
\resizebox{\textwidth}{!}{
\begin{tabular}{ll|cccccccccccc}
    \toprule
    &
    & \multicolumn{3}{c}{\textbf{GSM8K}} 
    & \multicolumn{3}{c}{\textbf{MATH500}} 
    & \multicolumn{3}{c}{\textbf{SVAMP}} 
    & \multicolumn{3}{c}{\textbf{Countdown}} \\
    \cmidrule(lr){3-5} \cmidrule(lr){6-8} \cmidrule(lr){9-11} \cmidrule(lr){12-14}
     & \textbf{Method / Seq Len} & 128 & 256 & 512 & 128 & 256 & 512 & 128 & 256 & 512 & 128 & 256 & 512 \\
    \midrule
    \multirow{1}{*}{\textbf{LLaDA-8B-Instruct}} 
    & baseline   & 68.5 & 76.3 & 78.2 & 27.4 & 33.4 & 35.8 & 84.0 & 83.3 & \textbf{84.7} & 20.3 & 21.5 & \textbf{16.4} \\
    \midrule
    \multirow{4}{*}{+ Temporal Voting} & Fixed Weighting & 68.0 & 73.4 & 78.3 & 26.6 & 30.8 & 34.2 & \textbf{87.0} & 84.3 & 84.3 & 22.7 & 18.8 & 11.3 \\
    \cmidrule(lr){2-14}
    & Linear Weighting & 70.0 & 78.0 & 78.8 & 28.0 & 34.4 & 34.6 & \textbf{87.0} & 84.3 & 84.3 & 24.2 & 21.9 & 16.0 \\
    \cmidrule(lr){2-14}
    & \multirow{2}{*}{Exp. Weighting} & \textbf{70.1} & \textbf{78.7} & \textbf{78.9} & \textbf{28.4} & \textbf{35.6} & \textbf{36.2} & 86.0 & \textbf{84.3} & \textbf{84.7} & \textbf{25.0} & \textbf{23.4} & \textbf{16.4} \\
    &  & \textcolor{green!70!black}{+1.6} & \textcolor{green!70!black}{+2.4} & \textcolor{green!70!black}{+0.7} & \textcolor{green!70!black}{+1.0} & \textcolor{green!70!black}{+2.2} & \textcolor{green!70!black}{+0.4} & \textcolor{green!70!black}{+2.0} & \textcolor{green!70!black}{+1.0} & \textcolor{gray}{+0.0} & \textcolor{green!70!black}{+4.7} & \textcolor{green!70!black}{+1.9} & \textcolor{gray}{+0.0} \\
    \midrule
    & \demphs{$\operatorname{EverPass} @ 1 \mid t$}  & \demphs{80.5} & \demphs{85.2} & \demphs{80.3} & \demphs{40.2} & \demphs{45.6} & \demphs{47.2} & \demphs{91.3} & \demphs{89.3} & \demphs{86.7} & \demphs{28.1} & \demphs{27.7} & \demphs{21.1} \\
    \midrule
    \midrule
    \multirow{1}{*}{\textbf{LLaDA-1.5}} 
    & baseline   & 69.8 & 79.4 & \textbf{81.1} & {29.0} & 32.4 & 35.4 & 85.3 & 86.3 & 83.3 & 21.5 & 21.1 & 20.7 \\
    \midrule
    \multirow{4}{*}{+ Temporal Voting} & Fixed Weighting & 68.8 & 75.7 & 80.3 & 27.3 & 30.8 & 34.6 & \textbf{87.3} & 85.3 & 84.0 & 23.4 & 22.3 & 18.8 \\
    \cmidrule(lr){2-14}
    & Linear Weighting & \textbf{71.0} & \textbf{79.8} & 81.0 & \textbf{29.2} & 32.8 & 35.8 & 86.0 & 87.0 & 84.0 & 24.2 & 23.4 & 19.1 \\
    \cmidrule(lr){2-14}
    & \multirow{2}{*}{Exp. Weighting} & {70.7} & \textbf{79.8} & \textbf{81.1} & {29.0} & \textbf{33.2} & \textbf{36.2} & 85.7 & \textbf{87.7} & \textbf{84.3} & \textbf{26.2} & \textbf{25.0} & \textbf{21.1} \\
    &  & \textcolor{green!70!black}{+0.9} & \textcolor{green!70!black}{+0.4} & \textcolor{gray}{+0.0} & \textcolor{gray}{+0.0} & \textcolor{green!70!black}{+0.8} & \textcolor{green!70!black}{+0.8} & \textcolor{green!70!black}{+0.4} & \textcolor{green!70!black}{+1.4} & \textcolor{green!70!black}{+1.0} & \textcolor{green!70!black}{+4.7} & \textcolor{green!70!black}{+3.9} & \textcolor{green!70!black}{+0.4} \\
    \midrule
    & \demphs{$\operatorname{EverPass} @ 1 \mid t$}  & \demphs{81.5} & \demphs{88.9} & \demphs{83.6} & \demphs{39.8} & \demphs{47.4} & \demphs{49.2} & \demphs{90.7} & \demphs{90.3} & \demphs{86.0} & \demphs{30.5} & \demphs{27.0} & \demphs{25.4} \\
    \bottomrule
\end{tabular}
}
\end{table}

\subsection{Implementation Details}
We use LLaDA-8B-Instruct~\citep{nie2025large} and LLaDA-1.5~\citep{zhu2025llada1.5}, evaluating on four math benchmarks: GSM8K \citep{cobbe2021training}, MATH500 \citep{hendrycksmath2021}, SVAMP \citep{patel2021nlp}, and Countdown \citep{tinyzero}. Following d1~\citep{zhao2025d1}, we report performance under different output lengths. Temporal self-consistency voting applies exponential weights to sampling steps. For post-training, LLaDA-8B-Instruct undergoes supervised fine-tuning (SFT) on s1K~\citep{muennighoff2025s1} for 20 epochs with 4,096 token sequences, followed by reinforcement fine-tuning (RFT). 
% LLaDA-1.5 skips SFT, as it typically reduces performance due to prior extensive post-training. 
For LLaDA-1.5, we omit SFT because it often results in performance degradation, likely due to the model having already undergone sophisticated post-training.
All training uses 8 H800 GPUs. Further details are in \cref{app:implementation_details}.

\subsection{\methodnamevoting}
\label{subsec:exp_vote}

\textbf{Voting Strategies.}
We apply weighted voting across denoising steps using three schemes: fixed, linear, and exponential.  
Each uses a weighting function $f(t)$, where $t$ is the current diffusion step. 
The fixed scheme assigns equal weight to all steps with $f(t) = 1$.
The linear weighting takes the form $f(t)=1-t/T$, and exponential uses $f(t)=\exp(\alpha(1-t/T))$ with $\alpha=5$.
Both linear and exponential schemes prioritize early diffusion time steps, \ie, latter sampling steps.

\begin{figure*}
    \centering
    \includegraphics[width=\textwidth]{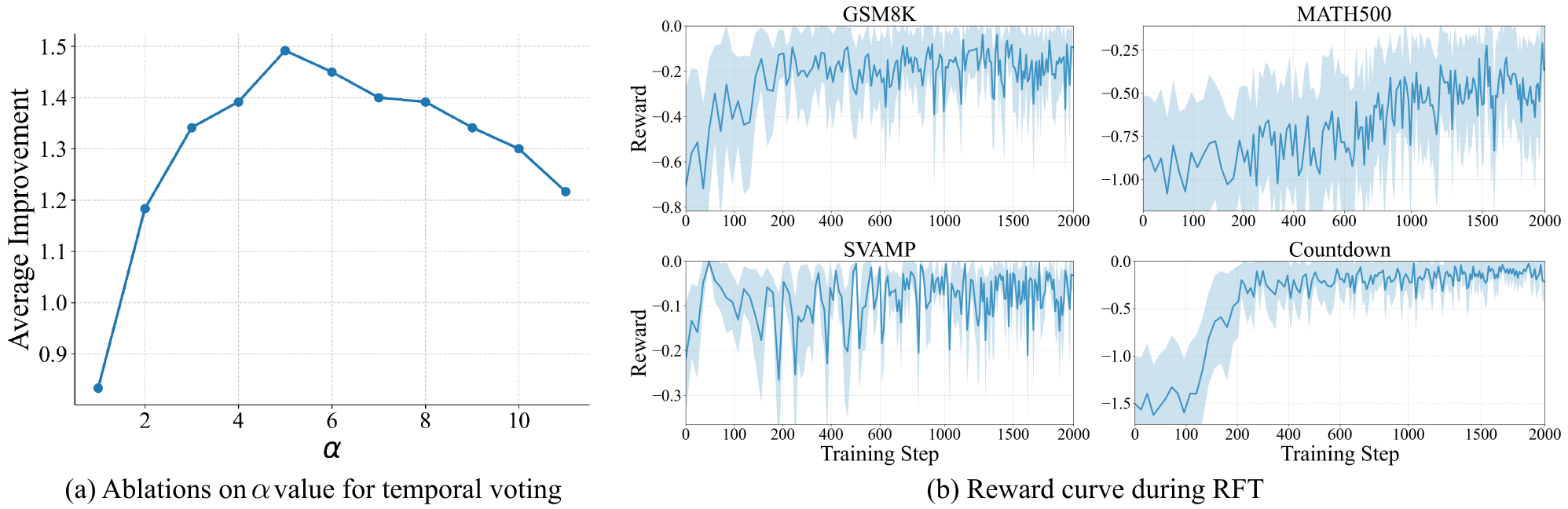}
    \caption{
        (a) Ablations on $\alpha$ value selection in temporal voting with exponential weighting.
        (b) Negative temporal semantic entropy reward curve during reinforcement fine-tuning.
    }
    \label{fig:reward}
\end{figure*}

\textbf{Ablations on Voting Strategies.}
As shown in \cref{tab:main_voting}, linear and exponential weighting improve inference performance, with exponential yielding the largest gains, \eg, LLaDA-8B-Instruct improves by 1.6\%, 1.2\%, 1.0\%, and 2.2\% on GSM8K, MATH500, SVAMP, and Countdown, respectively.  
Fixed weighting performs slightly worse, likely because equal weights amplify inaccurate early predictions. We therefore adopt exponential weighting by default.
We futher ablate the exponential hyperparameter $\alpha$. As shown in \cref{fig:reward}b, $\alpha$ values ranging from 1 to 11 consistently improve accuracy, peaking at $\alpha=5$ with an average gain of 1.5\%. Thus, we set $\alpha=5$ by default.

\subsection{\methodnamerft}

\begin{table}[t] 
\centering
\caption{
\textbf{Performance of reinforcement fine-tuning.}
Unlike d1~\citep{zhao2025d1}, which requires ground-truth answers, our method uses negative Temporal Semantic Entropy (TSE) as the reward without labels. Combining TSE with accuracy-based rewards yields further gains across benchmarks. 
\textcolor{green!70!black}{Green numbers} denote improvements over baseline.
} 
\label{table:main_rl} 
\vspace{2pt}
\resizebox{\textwidth}{!}{
\begin{tabular}{ll|cccccccccccc}
    \toprule
    &
    & \multicolumn{3}{c}{\textbf{GSM8K}} 
    & \multicolumn{3}{c}{\textbf{MATH500}} 
    & \multicolumn{3}{c}{\textbf{SVAMP}} 
    & \multicolumn{3}{c}{\textbf{Countdown}} \\
    \cmidrule(lr){3-5} \cmidrule(lr){6-8} \cmidrule(lr){9-11} \cmidrule(lr){12-14}
     & \textbf{Method / Seq Len} & 128 & 256 & 512 & 128 & 256 & 512 & 128 & 256 & 512 & 128 & 256 & 512 \\
    \midrule
    \multirow{1}{*}{\textbf{LLaDA}} 
    & baseline   & 70.2 & 78.7 & 80.1 & 23.8 & 34.4 & 36.8 & 81.7 & 83.3 & 82.7 & 21.5 & 19.9 & 21.5 \\
    \midrule
    \multirow{5}{*}{+ RFT}
    & accuracy reward (d1) & 71.7 & 78.3 & {82.3} & {31.0} & \textbf{36.0} & {40.4} & \textbf{85.7} & 88.0 & {88.7} & 34.8 & 35.5 & 37.9 \\
    \cmidrule(lr){2-14}
    & \multirow{2}{*}{negative TSE reward (ours)} & \textbf{72.2} & {78.8} & 80.2 & 30.6 & 34.6 & 38.0 & 84.3 & {89.0} & {88.7} & {38.6} & \textbf{53.5} & {44.9} \\
    &  & \textcolor{green!70!black}{+2.0} & \textcolor{green!70!black}{+0.1} & \textcolor{green!70!black}{+0.1} & \textcolor{green!70!black}{+6.8} & \textcolor{green!70!black}{+0.2} & \textcolor{green!70!black}{+1.2} & \textcolor{green!70!black}{+2.6} & \textcolor{green!70!black}{+5.7} & \textcolor{green!70!black}{+6.0} & \textcolor{green!70!black}{+17.1} & \textcolor{green!70!black}{+33.6} & \textcolor{green!70!black}{+23.4} \\
    \cmidrule(lr){2-14}
    & \multirow{2}{*}{combining both (ours)} & 72.1 & \textbf{80.0} & \textbf{83.0} & \textbf{31.2} & 35.4 & \textbf{41.4} & 85.0 & \textbf{90.3} & \textbf{92.3} & \textbf{41.5} & 42.6 & \textbf{54.7}  \\
    &  & \textcolor{green!70!black}{+1.9} & \textcolor{green!70!black}{+1.3} & \textcolor{green!70!black}{+2.9} & \textcolor{green!70!black}{+7.4} & \textcolor{green!70!black}{+1.0} & \textcolor{green!70!black}{+4.6} & \textcolor{green!70!black}{+3.3} & \textcolor{green!70!black}{+7.0} & \textcolor{green!70!black}{+9.6} & \textcolor{green!70!black}{+20.0} & \textcolor{green!70!black}{+22.7} & \textcolor{green!70!black}{+33.2} \\
    \midrule
    \midrule
    \multirow{1}{*}{\textbf{LLaDA-1.5}} 
    & baseline   & 69.8 & 79.4 & 81.1 & 29.0 & 32.4 & 35.4 & 85.3 & 86.3 & 83.3 & 21.5 & 21.1 & 20.7 \\
    \midrule
    \multirow{5}{*}{+ RFT} & 
    accuracy reward (d1)  & 73.0 & 78.9 & 83.1 & 29.8 & \textbf{36.2} & 40.2 & 84.7 & 89.3 & 88.0 & 38.7 & 29.7 & 39.1 \\
    \cmidrule(lr){2-14}
    & \multirow{2}{*}{negative TSE reward (ours)} & 72.0 & \textbf{80.8} & 82.6 & \textbf{30.2} & 35.0 & 40.0 & \textbf{86.3} & 88.3 & 87.3 & \textbf{50.0} & \textbf{55.9} & 53.1 \\
    &  & \textcolor{green!70!black}{+2.2} & \textcolor{green!70!black}{+1.4} & \textcolor{green!70!black}{+1.5} & \textcolor{green!70!black}{+1.2} & \textcolor{green!70!black}{+2.6} & \textcolor{green!70!black}{+4.6} & \textcolor{green!70!black}{+1.0} & \textcolor{green!70!black}{+2.0} & \textcolor{green!70!black}{+4.0} & \textcolor{green!70!black}{+28.5} & \textcolor{green!70!black}{+34.8} & \textcolor{green!70!black}{+32.4} \\
    \cmidrule(lr){2-14}
    & \multirow{2}{*}{combining both (ours)}
    & \textbf{73.2} & {80.5} & \textbf{84.0} & 29.6 & 35.4 & \textbf{41.4} & \textbf{86.3} & \textbf{90.3} & \textbf{89.0} & {44.5} & {46.9} & \textbf{63.3} \\
    &  & \textcolor{green!70!black}{+3.4} & \textcolor{green!70!black}{+1.1} & \textcolor{green!70!black}{+2.9} & \textcolor{green!70!black}{+0.6} & \textcolor{green!70!black}{+3.0} & \textcolor{green!70!black}{+6.0} & \textcolor{green!70!black}{+1.0} & \textcolor{green!70!black}{+4.0} & \textcolor{green!70!black}{+5.7} & \textcolor{green!70!black}{+23.0} & \textcolor{green!70!black}{+25.8} & \textcolor{green!70!black}{+42.6} \\
    \bottomrule
\end{tabular}
}
\end{table}

\textbf{Main Results.}
\cref{table:main_rl} reports results of incorporating temporal consistency into RFT.
We have the following observations.
\textbf{(1)} Using TSE reward alone consistently improves performance across lengths and datasets.  
\textbf{(2)} TSE reward matches or surpasses accuracy reward despite not using ground truth, \eg, on Countdown, LLaDA-8B-Instruct improves 24.7\% vs. 15.1\% with d1.
\textbf{(3)} Combining TSE with accuracy reward further boosts results, with absolute gains of 0.9\% (GSM8K), 0.2\% (MATH500), 1.7\% (SVAMP), and 10.2\% (Countdown) over d1.
Overall, our method achieves average improvements of 2.0\%, 4.3\%, 6.6\%, and 25.3\% over the SFT baseline, confirming the benefit of encouraging temporal consistency.

\textbf{Training Dynamics.}
We visualize the reward curves during training using LLaDA-8B-Instruct as an example, as shown in \cref{fig:reward}. The curves demonstrate a consistent upward trend in rewards across different datasets as training progresses, indicating effective learning and stable optimization.

\begin{figure*}
    \centering
    \includegraphics[width=\textwidth]{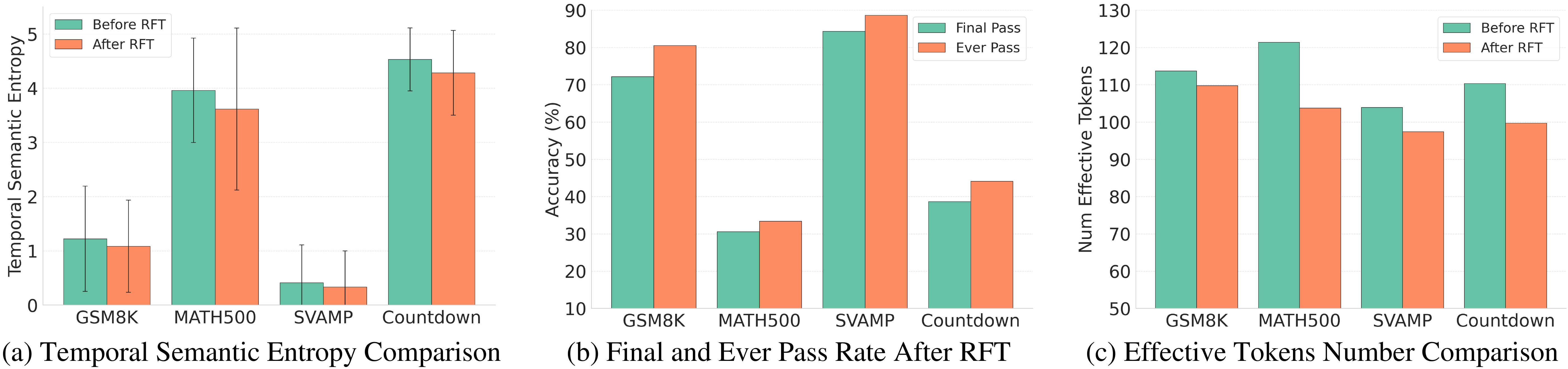}
    \caption{
    \textbf{Model attributes after reinforcement finetuning (RFT).}
    (a) Temporal semantic entropy decreases after RFT, showing improved semantic consistency in outputs.  
    (b) The \metricname remains above the final pass rate, leaving room for further gains.  
    (c) Effective tokens per generation drop after RFT, yielding more concise outputs.
    }
    \label{fig:rft_properties}
\end{figure*}

\textbf{Model Attributes After RFT.}
We analyze LLaDA-8B-Instruct fine-tuned with the negative TSE reward, with generation length 128, evaluating its behavior across several dimensions: TSE, the \metricname, and the number of effective tokens (defined as the average count of non-padding, non-EOS tokens per generation).
We have the following observations.
\textbf{(1)} As shown in \cref{fig:rft_properties}a, temporal semantic entropy consistently decreases across various datasets after RFT, reflecting enhanced temporal consistency in the model’s outputs—an anticipated result of reinforcement learning; 
\textbf{(2)} \cref{fig:rft_properties}b demonstrates that the \metricname remains higher than the final pass rate after RFT, suggesting there is still potential for further improvement.
\textbf{(3)} In \cref{fig:rft_properties}c, effective tokens decline after RFT, implying more concise outputs. We suspect shorter generations may reduce temporal oscillations, though this requires further study.
Further analysis of model attributes after RFT and a detailed discussion of limitations are provided in \cref{app:subsec_vote_after_rl} and \cref{app:sec_limitations}.

\section{Conclusion}

This work uncovers a critical yet overlooked aspect of diffusion large language models: their rich temporal dynamics. By identifying temporal oscillation as a recurring issue in dLLM decoding, we challenge the convention of relying solely on final-step predictions. Our proposed methods—\methodnamevoting and \methodnamerft—demonstrate that intermediate predictions are not noise, but signal. These strategies improve accuracy and stability without requiring additional inference passes or ground-truth supervision. Through extensive experiments, we show that temporal consistency is not just a desirable property—it’s a powerful lever for performance. We hope that this study inspires future research to \textit{treat 
intermediate denoising time steps
not as a nuisance, but as a feature in diffusion-based text generation}.

\section*{Acknowledgment}
We would like to thank Muzhi Zhu, Canyu Zhao, and Linhao Zhong at Zhejiang University for their valuable discussions and insightful feedback.

\section*{Ethics Statement}
This work does not involve human subjects, personal data, or sensitive information. All datasets used in our experiments (GSM8K, MATH500, SVAMP, and Countdown) are publicly available benchmark datasets designed for evaluating mathematical reasoning in large language models. We strictly adhered to ethical research practices and did not conduct any data collection that could raise privacy, security, or fairness concerns. Our methods—Temporal Self-Consistency Voting and Temporal Consistency Reinforcement—focus on improving the robustness and accuracy of diffusion language models, without introducing risks of harmful applications. To the best of our knowledge, this research complies with the ICLR Code of Ethics and poses no foreseeable ethical concerns.

\section*{Reproducibility Statement}
We have made extensive efforts to ensure the reproducibility of our work. Detailed dataset descriptions are provided in \cref{app:subsec_dataseat}, and training configurations and hyperparameters are reported in \cref{app:subsec_train}. The sampling and evaluation procedures are outlined in \cref{app:subsec_eval}. 
% Extended experimental results, ablation studies, and failure case analyses are included in Appendices C–E. 
To further promote transparency, we provide mathematical formulations and thorough descriptions of our proposed algorithms—Temporal Self-Consistency Voting and Temporal Consistency Reinforcement—directly in the main paper. 
Upon acceptance, we will release our models, together with training and inference code, to facilitate replication and further research.

{
\bibliographystyle{iclr2026_conference}
\bibliography{ref}

\begin{thebibliography}{59}
\providecommand{\natexlab}[1]{#1}
\providecommand{\url}[1]{\texttt{#1}}
\expandafter\ifx\csname urlstyle\endcsname\relax
  \providecommand{\doi}[1]{doi: #1}\else
  \providecommand{\doi}{doi: \begingroup \urlstyle{rm}\Url}\fi

\bibitem[Agarwal et~al.(2025)Agarwal, Zhang, Yuan, Han, and Peng]{agarwal2025unreasonable}
Shivam Agarwal, Zimin Zhang, Lifan Yuan, Jiawei Han, and Hao Peng.
\newblock The unreasonable effectiveness of entropy minimization in llm reasoning.
\newblock \emph{arXiv preprint arXiv:2505.15134}, 2025.

\bibitem[Arriola et~al.(2025)Arriola, Gokaslan, Chiu, Yang, Qi, Han, Sahoo, and Kuleshov]{arriola2025block}
Marianne Arriola, Aaron Gokaslan, Justin~T Chiu, Zhihan Yang, Zhixuan Qi, Jiaqi Han, Subham~Sekhar Sahoo, and Volodymyr Kuleshov.
\newblock Block diffusion: Interpolating between autoregressive and diffusion language models.
\newblock \emph{arXiv preprint arXiv:2503.09573}, 2025.

\bibitem[Austin et~al.(2021)Austin, Johnson, Ho, Tarlow, and Van Den~Berg]{austin2021structured}
Jacob Austin, Daniel~D Johnson, Jonathan Ho, Daniel Tarlow, and Rianne Van Den~Berg.
\newblock Structured denoising diffusion models in discrete state-spaces.
\newblock \emph{Advances in neural information processing systems}, 34:\penalty0 17981--17993, 2021.

\bibitem[Avdeyev et~al.(2023)Avdeyev, Shi, Tan, Dudnyk, and Zhou]{avdeyev2023dirichlet}
Pavel Avdeyev, Chenlai Shi, Yuhao Tan, Kseniia Dudnyk, and Jian Zhou.
\newblock Dirichlet diffusion score model for biological sequence generation.
\newblock In \emph{International Conference on Machine Learning}, pp.\  1276--1301. PMLR, 2023.

\bibitem[Chen et~al.(2021)Chen, Tworek, Jun, Yuan, Pinto, Kaplan, Edwards, Burda, Joseph, Brockman, et~al.]{chen2021evaluating}
Mark Chen, Jerry Tworek, Heewoo Jun, Qiming Yuan, Henrique Ponde De~Oliveira Pinto, Jared Kaplan, Harri Edwards, Yuri Burda, Nicholas Joseph, Greg Brockman, et~al.
\newblock Evaluating large language models trained on code.
\newblock \emph{arXiv preprint arXiv:2107.03374}, 2021.

\bibitem[Chen et~al.(2025)Chen, Chen, Wang, and Yang]{chen2025seed}
Minghan Chen, Guikun Chen, Wenguan Wang, and Yi~Yang.
\newblock Seed-grpo: Semantic entropy enhanced grpo for uncertainty-aware policy optimization.
\newblock \emph{arXiv preprint arXiv:2505.12346}, 2025.

\bibitem[Cheng et~al.(2024)Cheng, Li, Peng, and Liu]{cheng2024categorical}
Chaoran Cheng, Jiahan Li, Jian Peng, and Ge~Liu.
\newblock Categorical flow matching on statistical manifolds.
\newblock \emph{Advances in Neural Information Processing Systems}, 37:\penalty0 54787--54819, 2024.

\bibitem[Cobbe et~al.(2021)Cobbe, Kosaraju, Bavarian, Chen, Jun, Kaiser, Plappert, Tworek, Hilton, Nakano, et~al.]{cobbe2021training}
Karl Cobbe, Vineet Kosaraju, Mohammad Bavarian, Mark Chen, Heewoo Jun, Lukasz Kaiser, Matthias Plappert, Jerry Tworek, Jacob Hilton, Reiichiro Nakano, et~al.
\newblock Training verifiers to solve math word problems.
\newblock \emph{arXiv preprint arXiv:2110.14168}, 2021.

\bibitem[Cui et~al.(2025)Cui, Zhang, Chen, Yuan, Wang, Zuo, Li, Fan, Chen, Chen, et~al.]{cui2025entropy}
Ganqu Cui, Yuchen Zhang, Jiacheng Chen, Lifan Yuan, Zhi Wang, Yuxin Zuo, Haozhan Li, Yuchen Fan, Huayu Chen, Weize Chen, et~al.
\newblock The entropy mechanism of reinforcement learning for reasoning language models.
\newblock \emph{arXiv preprint arXiv:2505.22617}, 2025.

\bibitem[Damani et~al.(2025)Damani, Puri, Slocum, Shenfeld, Choshen, Kim, and Andreas]{damani2025beyond}
Mehul Damani, Isha Puri, Stewart Slocum, Idan Shenfeld, Leshem Choshen, Yoon Kim, and Jacob Andreas.
\newblock Beyond binary rewards: Training lms to reason about their uncertainty.
\newblock \emph{arXiv preprint arXiv:2507.16806}, 2025.

\bibitem[Davis et~al.(2024)Davis, Kessler, Petrache, Ceylan, Bronstein, and Bose]{davis2024fisher}
Oscar Davis, Samuel Kessler, Mircea Petrache, Ismail Ceylan, Michael Bronstein, and Joey Bose.
\newblock Fisher flow matching for generative modeling over discrete data.
\newblock \emph{Advances in Neural Information Processing Systems}, 37:\penalty0 139054--139084, 2024.

\bibitem[Farquhar et~al.(2024)Farquhar, Kossen, Kuhn, and Gal]{farquhar2024detecting}
Sebastian Farquhar, Jannik Kossen, Lorenz Kuhn, and Yarin Gal.
\newblock Detecting hallucinations in large language models using semantic entropy.
\newblock \emph{Nature}, 630\penalty0 (8017):\penalty0 625--630, 2024.

\bibitem[Gneiting \& Raftery(2007)Gneiting and Raftery]{gneiting2007strictly}
Tilmann Gneiting and Adrian~E Raftery.
\newblock Strictly proper scoring rules, prediction, and estimation.
\newblock \emph{Journal of the American statistical Association}, 102\penalty0 (477):\penalty0 359--378, 2007.

\bibitem[Gong et~al.(2025)Gong, Zhang, Zheng, Gu, Jaitly, Kong, and Zhang]{gong2025diffucoder}
Shansan Gong, Ruixiang Zhang, Huangjie Zheng, Jiatao Gu, Navdeep Jaitly, Lingpeng Kong, and Yizhe Zhang.
\newblock Diffucoder: Understanding and improving masked diffusion models for code generation.
\newblock \emph{arXiv preprint arXiv:2506.20639}, 2025.

\bibitem[Gulrajani \& Hashimoto(2023)Gulrajani and Hashimoto]{gulrajani2023likelihood}
Ishaan Gulrajani and Tatsunori~B Hashimoto.
\newblock Likelihood-based diffusion language models.
\newblock \emph{Advances in Neural Information Processing Systems}, 36:\penalty0 16693--16715, 2023.

\bibitem[Guo et~al.(2025)Guo, Yang, Zhang, Song, Zhang, Xu, Zhu, Ma, Wang, Bi, et~al.]{guo2025deepseek}
Daya Guo, Dejian Yang, Haowei Zhang, Junxiao Song, Ruoyu Zhang, Runxin Xu, Qihao Zhu, Shirong Ma, Peiyi Wang, Xiao Bi, et~al.
\newblock Deepseek-r1: Incentivizing reasoning capability in llms via reinforcement learning.
\newblock \emph{arXiv preprint arXiv:2501.12948}, 2025.

\bibitem[Han et~al.(2022)Han, Kumar, and Tsvetkov]{han2022ssd}
Xiaochuang Han, Sachin Kumar, and Yulia Tsvetkov.
\newblock Ssd-lm: Semi-autoregressive simplex-based diffusion language model for text generation and modular control.
\newblock \emph{arXiv preprint arXiv:2210.17432}, 2022.

\bibitem[Hendrycks et~al.(2021)Hendrycks, Burns, Kadavath, Arora, Basart, Tang, Song, and Steinhardt]{hendrycksmath2021}
Dan Hendrycks, Collin Burns, Saurav Kadavath, Akul Arora, Steven Basart, Eric Tang, Dawn Song, and Jacob Steinhardt.
\newblock Measuring mathematical problem solving with the math dataset.
\newblock \emph{NeurIPS}, 2021.

\bibitem[Ho et~al.(2020)Ho, Jain, and Abbeel]{ho2020denoising}
Jonathan Ho, Ajay Jain, and Pieter Abbeel.
\newblock Denoising diffusion probabilistic models.
\newblock \emph{Advances in neural information processing systems}, 33:\penalty0 6840--6851, 2020.

\bibitem[Ho et~al.(2022)Ho, Salimans, Gritsenko, Chan, Norouzi, and Fleet]{ho2022video}
Jonathan Ho, Tim Salimans, Alexey Gritsenko, William Chan, Mohammad Norouzi, and David~J Fleet.
\newblock Video diffusion models.
\newblock \emph{Advances in neural information processing systems}, 35:\penalty0 8633--8646, 2022.

\bibitem[Hu et~al.(2022)Hu, Shen, Wallis, Allen-Zhu, Li, Wang, Wang, Chen, et~al.]{hu2022lora}
Edward~J Hu, Yelong Shen, Phillip Wallis, Zeyuan Allen-Zhu, Yuanzhi Li, Shean Wang, Lu~Wang, Weizhu Chen, et~al.
\newblock Lora: Low-rank adaptation of large language models.
\newblock \emph{ICLR}, 1\penalty0 (2):\penalty0 3, 2022.

\bibitem[Huang et~al.(2025)Huang, Jia, Zhai, Cao, Ye, Zhao, Xu, Hu, and Lin]{huang2025vision}
Wenxuan Huang, Bohan Jia, Zijie Zhai, Shaosheng Cao, Zheyu Ye, Fei Zhao, Zhe Xu, Yao Hu, and Shaohui Lin.
\newblock Vision-r1: Incentivizing reasoning capability in multimodal large language models.
\newblock \emph{arXiv preprint arXiv:2503.06749}, 2025.

\bibitem[Kim et~al.(2025)Kim, Shah, Kontonis, Kakade, and Chen]{kim2025train}
Jaeyeon Kim, Kulin Shah, Vasilis Kontonis, Sham Kakade, and Sitan Chen.
\newblock Train for the worst, plan for the best: Understanding token ordering in masked diffusions.
\newblock \emph{arXiv preprint arXiv:2502.06768}, 2025.

\bibitem[Kuhn et~al.(2023)Kuhn, Gal, and Farquhar]{kuhn2023semantic}
Lorenz Kuhn, Yarin Gal, and Sebastian Farquhar.
\newblock Semantic uncertainty: Linguistic invariances for uncertainty estimation in natural language generation.
\newblock \emph{arXiv preprint arXiv:2302.09664}, 2023.

\bibitem[Li et~al.(2022)Li, Thickstun, Gulrajani, Liang, and Hashimoto]{li2022diffusion}
Xiang Li, John Thickstun, Ishaan Gulrajani, Percy~S Liang, and Tatsunori~B Hashimoto.
\newblock Diffusion-lm improves controllable text generation.
\newblock \emph{Advances in neural information processing systems}, 35:\penalty0 4328--4343, 2022.

\bibitem[Li et~al.(2025)Li, Xu, Jiang, Ramasubramanian, Niu, Lin, Yue, and Poovendran]{li2025temporal}
Yuetai Li, Zhangchen Xu, Fengqing Jiang, Bhaskar Ramasubramanian, Luyao Niu, Bill~Yuchen Lin, Xiang Yue, and Radha Poovendran.
\newblock Temporal sampling for forgotten reasoning in llms.
\newblock \emph{arXiv preprint arXiv:2505.20196}, 2025.

\bibitem[Lightman et~al.(2023)Lightman, Kosaraju, Burda, Edwards, Baker, Lee, Leike, Schulman, Sutskever, and Cobbe]{lightman2023let}
Hunter Lightman, Vineet Kosaraju, Yuri Burda, Harrison Edwards, Bowen Baker, Teddy Lee, Jan Leike, John Schulman, Ilya Sutskever, and Karl Cobbe.
\newblock Let's verify step by step.
\newblock In \emph{The Twelfth International Conference on Learning Representations}, 2023.

\bibitem[Liu et~al.(2025{\natexlab{a}})Liu, Li, Fang, Xu, He, and Tan]{liu2025rethinking}
Yexiang Liu, Zekun Li, Zhi Fang, Nan Xu, Ran He, and Tieniu Tan.
\newblock Rethinking the role of prompting strategies in llm test-time scaling: A perspective of probability theory.
\newblock \emph{arXiv preprint arXiv:2505.10981}, 2025{\natexlab{a}}.

\bibitem[Liu et~al.(2025{\natexlab{b}})Liu, Chen, Li, Qi, Pang, Du, Lee, and Lin]{liu2025understanding}
Zichen Liu, Changyu Chen, Wenjun Li, Penghui Qi, Tianyu Pang, Chao Du, Wee~Sun Lee, and Min Lin.
\newblock Understanding r1-zero-like training: A critical perspective.
\newblock \emph{arXiv preprint arXiv:2503.20783}, 2025{\natexlab{b}}.

\bibitem[Lou et~al.(2023)Lou, Meng, and Ermon]{lou2023discrete}
Aaron Lou, Chenlin Meng, and Stefano Ermon.
\newblock Discrete diffusion modeling by estimating the ratios of the data distribution.
\newblock \emph{arXiv preprint arXiv:2310.16834}, 2023.

\bibitem[Madaan et~al.(2023)Madaan, Tandon, Gupta, Hallinan, Gao, Wiegreffe, Alon, Dziri, Prabhumoye, Yang, et~al.]{madaan2023self}
Aman Madaan, Niket Tandon, Prakhar Gupta, Skyler Hallinan, Luyu Gao, Sarah Wiegreffe, Uri Alon, Nouha Dziri, Shrimai Prabhumoye, Yiming Yang, et~al.
\newblock Self-refine: Iterative refinement with self-feedback.
\newblock \emph{Advances in Neural Information Processing Systems}, 36:\penalty0 46534--46594, 2023.

\bibitem[Muennighoff et~al.(2025)Muennighoff, Yang, Shi, Li, Fei-Fei, Hajishirzi, Zettlemoyer, Liang, Cand{\`e}s, and Hashimoto]{muennighoff2025s1}
Niklas Muennighoff, Zitong Yang, Weijia Shi, Xiang~Lisa Li, Li~Fei-Fei, Hannaneh Hajishirzi, Luke Zettlemoyer, Percy Liang, Emmanuel Cand{\`e}s, and Tatsunori Hashimoto.
\newblock s1: Simple test-time scaling.
\newblock \emph{arXiv preprint arXiv:2501.19393}, 2025.

\bibitem[Nie et~al.(2024)Nie, Zhu, Du, Pang, Liu, Zeng, Lin, and Li]{nie2024scaling}
Shen Nie, Fengqi Zhu, Chao Du, Tianyu Pang, Qian Liu, Guangtao Zeng, Min Lin, and Chongxuan Li.
\newblock Scaling up masked diffusion models on text.
\newblock \emph{arXiv preprint arXiv:2410.18514}, 2024.

\bibitem[Nie et~al.(2025)Nie, Zhu, You, Zhang, Ou, Hu, Zhou, Lin, Wen, and Li]{nie2025large}
Shen Nie, Fengqi Zhu, Zebin You, Xiaolu Zhang, Jingyang Ou, Jun Hu, Jun Zhou, Yankai Lin, Ji-Rong Wen, and Chongxuan Li.
\newblock Large language diffusion models.
\newblock \emph{arXiv preprint arXiv:2502.09992}, 2025.

\bibitem[Pan et~al.(2025)Pan, Zhang, Wang, Yuan, Peng, and Suhr]{tinyzero}
Jiayi Pan, Junjie Zhang, Xingyao Wang, Lifan Yuan, Hao Peng, and Alane Suhr.
\newblock Tinyzero.
\newblock https://github.com/Jiayi-Pan/TinyZero, 2025.
\newblock Accessed: 2025-01-24.

\bibitem[Patel et~al.(2021)Patel, Bhattamishra, and Goyal]{patel2021nlp}
Arkil Patel, Satwik Bhattamishra, and Navin Goyal.
\newblock Are nlp models really able to solve simple math word problems?
\newblock \emph{arXiv preprint arXiv:2103.07191}, 2021.

\bibitem[Prabhudesai et~al.(2025)Prabhudesai, Chen, Ippoliti, Fragkiadaki, Liu, and Pathak]{prabhudesai2025rent}
Mihir Prabhudesai, Lili Chen, Alex Ippoliti, Katerina Fragkiadaki, Hao Liu, and Deepak Pathak.
\newblock Maximizing confidence alone improves reasoning.
\newblock \emph{arXiv preprint arXiv:2505.22660}, 2025.

\bibitem[Qi et~al.(2025)Qi, Zhang, Yu, Wang, and Zhao]{qi2025vln}
Zhangyang Qi, Zhixiong Zhang, Yizhou Yu, Jiaqi Wang, and Hengshuang Zhao.
\newblock Vln-r1: Vision-language navigation via reinforcement fine-tuning.
\newblock \emph{arXiv preprint arXiv:2506.17221}, 2025.

\bibitem[Radford et~al.(2019)Radford, Wu, Child, Luan, Amodei, Sutskever, et~al.]{radford2019language}
Alec Radford, Jeffrey Wu, Rewon Child, David Luan, Dario Amodei, Ilya Sutskever, et~al.
\newblock Language models are unsupervised multitask learners.
\newblock \emph{OpenAI blog}, 1\penalty0 (8):\penalty0 9, 2019.

\bibitem[Sahoo et~al.(2024)Sahoo, Arriola, Schiff, Gokaslan, Marroquin, Chiu, Rush, and Kuleshov]{sahoo2024simple}
Subham Sahoo, Marianne Arriola, Yair Schiff, Aaron Gokaslan, Edgar Marroquin, Justin Chiu, Alexander Rush, and Volodymyr Kuleshov.
\newblock Simple and effective masked diffusion language models.
\newblock \emph{Advances in Neural Information Processing Systems}, 37:\penalty0 130136--130184, 2024.

\bibitem[Sahoo et~al.(2025)Sahoo, Yang, Akhauri, Liu, Singh, Cheng, Liu, Xing, Thickstun, and Vahdat]{sahoo2025esoteric}
Subham~Sekhar Sahoo, Zhihan Yang, Yash Akhauri, Johnna Liu, Deepansha Singh, Zhoujun Cheng, Zhengzhong Liu, Eric Xing, John Thickstun, and Arash Vahdat.
\newblock Esoteric language models.
\newblock \emph{arXiv preprint arXiv:2506.01928}, 2025.

\bibitem[Schulman et~al.(2017)Schulman, Wolski, Dhariwal, Radford, and Klimov]{schulman2017proximal}
John Schulman, Filip Wolski, Prafulla Dhariwal, Alec Radford, and Oleg Klimov.
\newblock Proximal policy optimization algorithms.
\newblock \emph{arXiv preprint arXiv:1707.06347}, 2017.

\bibitem[Shao et~al.(2024)Shao, Wang, Zhu, Xu, Song, Bi, Zhang, Zhang, Li, Wu, et~al.]{shao2024deepseekmath}
Zhihong Shao, Peiyi Wang, Qihao Zhu, Runxin Xu, Junxiao Song, Xiao Bi, Haowei Zhang, Mingchuan Zhang, YK~Li, Yang Wu, et~al.
\newblock Deepseekmath: Pushing the limits of mathematical reasoning in open language models.
\newblock \emph{arXiv preprint arXiv:2402.03300}, 2024.

\bibitem[Shen et~al.(2025)Shen, Liu, Li, Fang, Ma, Liao, Shen, Zhang, Zhao, Zhang, et~al.]{shen2025vlm}
Haozhan Shen, Peng Liu, Jingcheng Li, Chunxin Fang, Yibo Ma, Jiajia Liao, Qiaoli Shen, Zilun Zhang, Kangjia Zhao, Qianqian Zhang, et~al.
\newblock Vlm-r1: A stable and generalizable r1-style large vision-language model.
\newblock \emph{arXiv preprint arXiv:2504.07615}, 2025.

\bibitem[Shi et~al.(2024)Shi, Han, Wang, Doucet, and Titsias]{shi2024simplified}
Jiaxin Shi, Kehang Han, Zhe Wang, Arnaud Doucet, and Michalis~K. Titsias.
\newblock Simplified and generalized masked diffusion for discrete data.
\newblock In \emph{Advances in Neural Information Processing Systems}, 2024.

\bibitem[Snell et~al.(2024)Snell, Lee, Xu, and Kumar]{snell2024scaling}
Charlie Snell, Jaehoon Lee, Kelvin Xu, and Aviral Kumar.
\newblock Scaling llm test-time compute optimally can be more effective than scaling model parameters.
\newblock \emph{arXiv preprint arXiv:2408.03314}, 2024.

\bibitem[Song et~al.(2020)Song, Meng, and Ermon]{song2020denoising}
Jiaming Song, Chenlin Meng, and Stefano Ermon.
\newblock Denoising diffusion implicit models.
\newblock \emph{arXiv preprint arXiv:2010.02502}, 2020.

\bibitem[Stark et~al.(2024)Stark, Jing, Wang, Corso, Berger, Barzilay, and Jaakkola]{stark2024dirichlet}
Hannes Stark, Bowen Jing, Chenyu Wang, Gabriele Corso, Bonnie Berger, Regina Barzilay, and Tommi Jaakkola.
\newblock Dirichlet flow matching with applications to dna sequence design.
\newblock \emph{arXiv preprint arXiv:2402.05841}, 2024.

\bibitem[Wang et~al.(2025)Wang, Schiff, Sahoo, and Kuleshov]{wang2025remasking}
Guanghan Wang, Yair Schiff, Subham~Sekhar Sahoo, and Volodymyr Kuleshov.
\newblock Remasking discrete diffusion models with inference-time scaling.
\newblock \emph{arXiv preprint arXiv:2503.00307}, 2025.

\bibitem[Wang et~al.(2022)Wang, Wei, Schuurmans, Le, Chi, Narang, Chowdhery, and Zhou]{wang2022self}
Xuezhi Wang, Jason Wei, Dale Schuurmans, Quoc Le, Ed~Chi, Sharan Narang, Aakanksha Chowdhery, and Denny Zhou.
\newblock Self-consistency improves chain of thought reasoning in language models.
\newblock \emph{arXiv preprint arXiv:2203.11171}, 2022.

\bibitem[Wei et~al.(2022)Wei, Wang, Schuurmans, Bosma, Xia, Chi, Le, Zhou, et~al.]{wei2022chain}
Jason Wei, Xuezhi Wang, Dale Schuurmans, Maarten Bosma, Fei Xia, Ed~Chi, Quoc~V Le, Denny Zhou, et~al.
\newblock Chain-of-thought prompting elicits reasoning in large language models.
\newblock \emph{Advances in neural information processing systems}, 35:\penalty0 24824--24837, 2022.

\bibitem[Yang et~al.(2025)Yang, Tian, Li, Zhang, Shen, Tong, and Wang]{yang2025mmada}
Ling Yang, Ye~Tian, Bowen Li, Xinchen Zhang, Ke~Shen, Yunhai Tong, and Mengdi Wang.
\newblock Mmada: Multimodal large diffusion language models.
\newblock \emph{arXiv preprint arXiv:2505.15809}, 2025.

\bibitem[Yao et~al.(2023)Yao, Yu, Zhao, Shafran, Griffiths, Cao, and Narasimhan]{yao2023tree}
Shunyu Yao, Dian Yu, Jeffrey Zhao, Izhak Shafran, Tom Griffiths, Yuan Cao, and Karthik Narasimhan.
\newblock Tree of thoughts: Deliberate problem solving with large language models.
\newblock \emph{Advances in neural information processing systems}, 36:\penalty0 11809--11822, 2023.

\bibitem[Ye et~al.(2025)Ye, Xie, Zheng, Gao, Wu, Jiang, Li, and Kong]{dream2025}
Jiacheng Ye, Zhihui Xie, Lin Zheng, Jiahui Gao, Zirui Wu, Xin Jiang, Zhenguo Li, and Lingpeng Kong.
\newblock Dream 7b, 2025.
\newblock URL \url{https://hkunlp.github.io/blog/2025/dream}.

\bibitem[Yu et~al.(2025)Yu, Zhang, Zhu, Yuan, Zuo, Yue, Dai, Fan, Liu, Liu, et~al.]{yu2025dapo}
Qiying Yu, Zheng Zhang, Ruofei Zhu, Yufeng Yuan, Xiaochen Zuo, Yu~Yue, Weinan Dai, Tiantian Fan, Gaohong Liu, Lingjun Liu, et~al.
\newblock Dapo: An open-source llm reinforcement learning system at scale.
\newblock \emph{arXiv preprint arXiv:2503.14476}, 2025.

\bibitem[Zhang et~al.(2025)Zhang, Wu, Zhang, Zhao, and Bian]{zhang2025right}
Qingyang Zhang, Haitao Wu, Changqing Zhang, Peilin Zhao, and Yatao Bian.
\newblock Right question is already half the answer: Fully unsupervised llm reasoning incentivization.
\newblock \emph{arXiv preprint arXiv:2504.05812}, 2025.

\bibitem[Zhao et~al.(2025)Zhao, Gupta, Zheng, and Grover]{zhao2025d1}
Siyan Zhao, Devaansh Gupta, Qinqing Zheng, and Aditya Grover.
\newblock d1: Scaling reasoning in diffusion large language models via reinforcement learning.
\newblock \emph{arXiv preprint arXiv:2504.12216}, 2025.

\bibitem[Zhong et~al.(2025)Zhong, Zhu, Du, Huang, Zhao, Liu, Wang, Chen, and Shen]{zhong2025omni}
Hao Zhong, Muzhi Zhu, Zongze Du, Zheng Huang, Canyu Zhao, Mingyu Liu, Wen Wang, Hao Chen, and Chunhua Shen.
\newblock Omni-r1: Reinforcement learning for omnimodal reasoning via two-system collaboration.
\newblock \emph{arXiv preprint arXiv:2505.20256}, 2025.

\bibitem[Zhu et~al.(2025)Zhu, Wang, Nie, Zhang, Wu, Hu, Zhou, Chen, Lin, Wen, et~al.]{zhu2025llada1.5}
Fengqi Zhu, Rongzhen Wang, Shen Nie, Xiaolu Zhang, Chunwei Wu, Jun Hu, Jun Zhou, Jianfei Chen, Yankai Lin, Ji-Rong Wen, et~al.
\newblock Llada 1.5: Variance-reduced preference optimization for large language diffusion models.
\newblock \emph{arXiv preprint arXiv:2505.19223}, 2025.

\end{thebibliography}
}

\clearpage
\appendix
\renewcommand\thesection{\Alph{section}}
\renewcommand\thefigure{S\arabic{figure}}
\renewcommand\thetable{S\arabic{table}}
\renewcommand\theequation{S\arabic{equation}}
\setcounter{figure}{0}
\setcounter{table}{0}
\setcounter{equation}{0}

% \clearpage
\section*{Appendix}

\section*{LLM Usage}
\label{app:llm_usage}

In this section, we clarify the role of large language models (LLMs) in preparing this work. The model was used exclusively for language polishing, such as refining grammar, style, and readability, without contributing to the research design, analysis, or conclusions.

\section{Appendix Overview}

This appendix provides additional implementation details, empirical analysis, and extended results to supplement the main paper. It is organized as follows:

\begin{itemize}
    \item \textbf{\cref{app:llm_usage}: LLM Usage} \\
    Clarify the assistance from large language models (LLMs).
    
    \item \textbf{\cref{app:implementation_details}: More Implementation Details} \\
    Provides further implementation information, including:
    \begin{itemize}
        \item \cref{app:subsec_dataseat}: Detailed descriptions of the datasets used
        \item \cref{app:subsec_train}: Training configurations and hyperparameters
        \item \cref{app:subsec_eval}: Sampling strategies and evaluation procedures
    \end{itemize}

    \item \textbf{\cref{app:more_analysis}: More Analysis} \\
    Presents extended analyses, including:
    \begin{itemize}
        \item \cref{app:subsec_math_and_svamp}: Accuracy and entropy analysis on the MATH500 and SVAMP datasets
        \item \cref{app:subsec_tse_vary_len}: Temporal semantic entropy across varying generated lengths
        \item \cref{app:subsec_block_entropy}: Block-level token entropy analysis
    \end{itemize}

    \item \textbf{\cref{app:sec_limitations}: Limitations} \\
    Discuss limitations and analyze failure cases, including:
    \begin{itemize}
        \item \cref{app:subsec_limitations}: Discuss potential limitations of our method
        \item \cref{app:subsec_failure}: Analysis of failure cases on the Sudoku dataset
    \end{itemize}

    \item \textbf{\cref{app:more_exp}: More Experimental Results} \\
    Includes additional experimental findings, such as:
    \begin{itemize}
        \item \cref{app:subsec_joint_train}: Training a unified model across multiple tasks
        \item \cref{app:subsec_scoring_rules}: Ablation studies on different scoring rules for combining TSE with accuracy reward
        \item \cref{app:subsec_vote_after_rl}: Performance of temporal self-consistency voting on reinforcement fine-tuned models
        \item \cref{app:subsec_examples_oscillation}: Detailed examples illustrating temporal oscillation
    \end{itemize}
    
    \item \textbf{\cref{app:change_log}: Change Log} \\
    Summarizes the main modifications. 
\end{itemize}

\section{LLM Usage}
\label{app:llm_usage}

In this section, we clarify the role of large language models (LLMs) in preparing this work. The model was used exclusively for language polishing, such as refining grammar, style, and readability, without contributing to the research design, analysis, or conclusions.

\section{More Implementation Details}
\label{app:implementation_details}

\subsection{Datasets}
\label{app:subsec_dataseat}
We provided detailed descriptions of the datasets as follows:
\begin{itemize}
\item GSM8K \citep{cobbe2021training} comprises 8.5K linguistically diverse grade school math word problems (7.5K training, 1K test), solvable by bright middle school students via 2–8 steps of basic arithmetic, suited for multi-step mathematical reasoning.
\item MATH500 \citep{lightman2023let} is a curated subset of 500 problems selected from the broader MATH dataset~\citep{hendrycksmath2021}, featuring high-school-level competition math problems.
\item SVAMP \citep{patel2021nlp} serves as a benchmark for elementary-level Math Word Problems (MWPs), where each MWP is a short natural language narrative describing a scenario and asking questions about unknown quantities. 
\item Countdown \citep{tinyzero} involves a combinatorial arithmetic game with three numbers, requiring models to reach target numbers using basic arithmetic operations on a given set of numbers.
\end{itemize}

\subsection{Training}
\label{app:subsec_train}
During reinforcement fine-tuning, we train our model using sequences of 256 tokens, with a batch size of 6 per GPU and gradient accumulation over 2 steps.
Low-Rank Adaptation (LoRA)~\citep{hu2022lora} is applied with a rank of 128 and a scaling factor of 64.
During reward computation, answers are parsed from generated text sequences for semantic clustering. 
When answer parsing fails due to an inaccurate format, we simply discard the answer for temporal semantic entropy computation. 
Moreover, since answers generated in the first half of the sampling steps tend to be rough and less reliable, we exclude them from consideration. 
Only answers from the second half of the sampling steps are used to calculate the temporal semantic entropy.

\subsection{Sampling and Evaluation}
\label{app:subsec_eval}
During sampling, we adopt the semi-autoregressive sampling approach following LLaDA~\citep{nie2025large}.
Specifically, the sequence is split into multiple blocks, which are generated in a left-to-right manner. For each individual block, we employ the low-confidence remasking strategy during the sampling process.
Following the practice in d1~\citep{zhao2025d1}, we evaluate the model every 100 steps, starting from step 600 to  8,000 steps, and report the best results.

\section{More Analysis}
\label{app:more_analysis}

\subsection{Analysis on MATH500 and SVAMP datasets}
\label{app:subsec_math_and_svamp}

\begin{figure}[t]
    \centering
    \includegraphics[width=\textwidth]{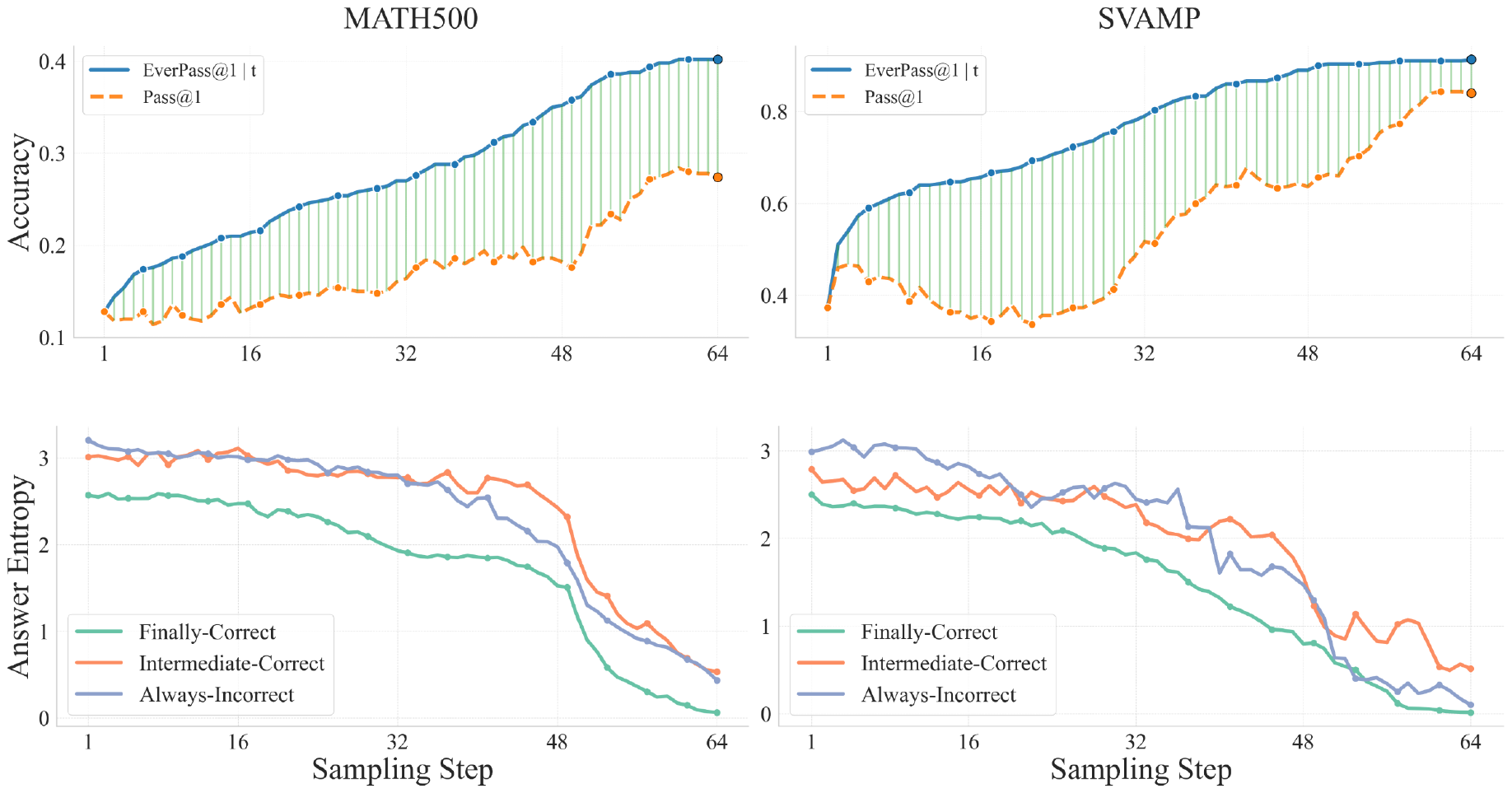}
    \caption{
        \textbf{Top Row:}
        $\operatorname{Pass} @ 1$ and $\operatorname{Pass} @ 1 \mid t$ for MATH500 and SVAMP are provided as supplementary results to \cref{fig:accuracy_analysis}, using the same experimental settings.
        A noticeable gap similar to that observed in GSM8K and Countdown between $\operatorname{Pass} @ 1$ and $\operatorname{Pass} @ 1 \mid t$ is present across all sampling steps.
        \textbf{Bottom Row:}
        Answer Entropy for MATH500 and SVAMP are provided as supplementary results to \cref{fig:entropy_analysis}, using the same experimental setups.
        MATH500 and SVAMP exhibit answer entropy patterns similar to those of GSM8K and Countdown.
    }
    \label{fig:app_analysis}
\end{figure}

\paragraph{Accuracy Analysis.}
As shown in the first row in \cref{fig:app_analysis}, MATH500 and SVAMP exhibit a similar pattern to that observed \cref{fig:accuracy_analysis} in \cref{subsec:observations_analy}, where a noticeable gap emerges between $\operatorname{Pass} @ 1$ and $\operatorname{Pass} @ 1 \mid t$.
In MATH500 and SVAMP, correct answers start to appear early in the sampling process(near 4\% and 39\% at first step, respectively), and continue to improve over subsequent iterations.
Interestingly, on the SVAMP dataset, a distinct pattern emerged in the model’s 
$\operatorname{Pass} @ 1$ accuracy across different sampling steps. 
Between steps 3 and 20, performance declined noticeably, followed by a recovery after step 20. 
This distinctive fluctuation trajectory represents another manifestation of the temporal oscillation phenomenon.

\paragraph{Entropy Analysis.}
As shown in the second row in \cref{fig:app_analysis}, for MATH500 and SVAMP, the answer entropy of \textit{Finally-Correct} questions remains the lowest throughout the sampling process.
\textit{Intermediate-Correct} questions consistently exhibit lower entropy in the early sampling steps compared to Always-Incorrect ones, a pattern observed across all four datasets.
However, unlike GSM8K, MATH500, and Countdown, where the final entropy of \textit{Intermediate-Correct} and \textit{Always-Incorrect} questions is similar, SVAMP displays relatively high entropy in the final sampling step for \textit{Intermediate-Correct} questions.

\subsection{Temporal Semantic Entropy over Varying Generated Length}
\label{app:subsec_tse_vary_len}

\begin{figure}[t]
    \centering
    \includegraphics[width=\linewidth]{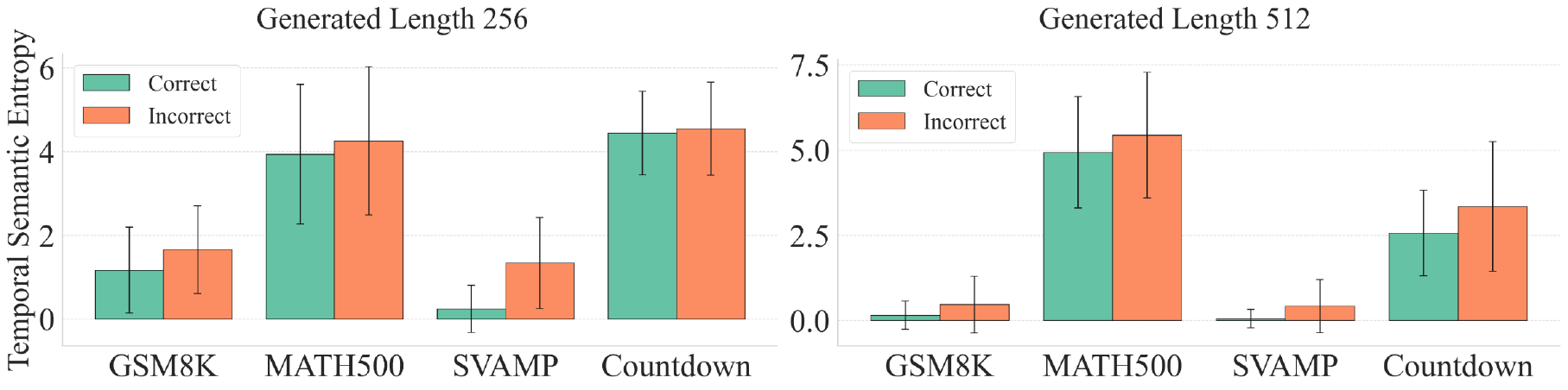}
    \caption{
        \textbf{Temporal semantic entropy with varying generation lengths.}
        Average temporal semantic entropy across datasets with generation lengths of 256 and 512, showing consistent patterns with \cref{fig:temporal_entropy}. 
        Higher entropy generally correlates with lower accuracy. 
    }
    \label{fig:app_temporal_entropy}
\end{figure}

To further validate the generalizability of our findings regarding temporal semantic entropy, we extended the experiments beyond those presented in \cref{fig:temporal_entropy}, which used a generation length of 128 and 64 diffusion steps. Specifically, we tested generation lengths of 256 and 512, with diffusion steps set to half the generation length.

As shown in \cref{fig:app_temporal_entropy}, the pattern of temporal semantic entropy observed here aligns with the conclusions drawn in \cref{subsec:observations_analy}: questions that ultimately receive incorrect answers consistently exhibit relatively high temporal semantic entropy throughout the sampling process. This high entropy reflects greater instability and uncertainty in the model’s intermediate predictions.

\subsection{Analysis on the Average Entropy in Blocks}
\label{app:subsec_block_entropy}

\begin{figure}
    \centering
    \includegraphics[width=\linewidth]{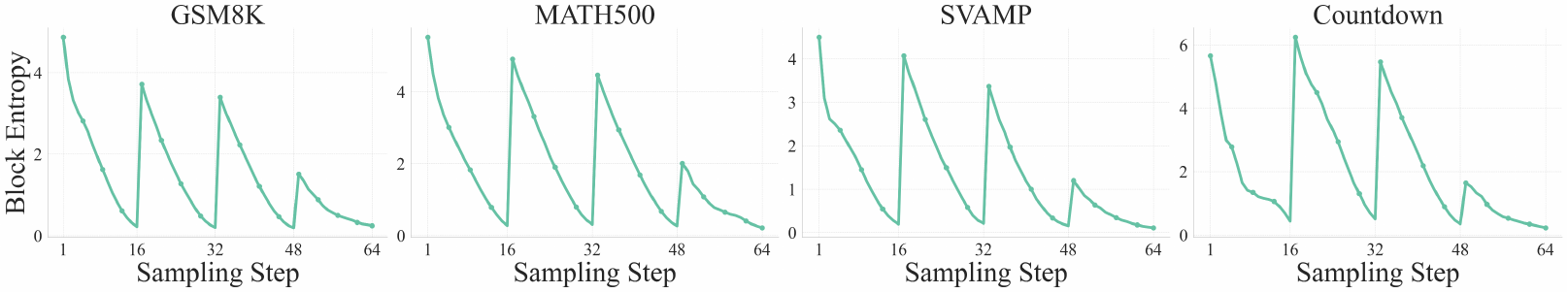}
    \caption{
        \textbf{Block-level entropy dynamics under semi-autoregressive sampling.} 
        During sampling, sequences are partitioned into fixed-length blocks, processed in a left-to-right order, with remasking and unmasking operations restricted to the current block.
        Average token entropy within a block decreases with more sampling steps. 
        A sharp entropy spike occurs when shifting to a new block, likely due to simultaneous decoding of multiple masked tokens increasing initial uncertainty.
    }
    \label{fig:app_block_entropy}
\end{figure}

In addition to the token-level entropy considered in \cref{subsec:observations_analy} and the answer-level entropy considered in temporal semantic entropy, we additionally introduce a block-level entropy. This is motivated by the fact that \dllm typically adopts a semi-autoregressive sampling strategy. 
In this sampling strategy, the entire generated sequence is divided into multiple fixed-length blocks, with each block allocated a specific number of sampling steps. During these steps, remasking and unmasking operations are performed exclusively on the current block. Starting with the first block, the process steps to the next block only after all tokens in the current block have been unmasked.  

We denote the current block as containing tokens indexed from a start $s$  to an end $e-1$ (inclusive), such that the block spans indices $i \in \{s, s+1, \ldots, e-1\}$. The block entropy is then calculated as:
\begin{equation*}
    H_{\text{block}} = \frac{1}{e - s} \sum_{i=s}^{e-1} H(i)
\end{equation*}
where $H(i)$ represents the entropy of the $i$-th generated token, and $e - s$ denotes the total number of tokens in the block.

As illustrated in \cref{fig:app_block_entropy}, the average token entropy within a single block exhibits a consistent downward trend as sampling steps accumulate. This pattern is intuitive: as more tokens in the current block are decoded, they collectively form a richer contextual foundation, thereby mitigating the model’s uncertainty about subsequent tokens in the block.
Interestingly, when decoding shifts to a new block, the entropy rises sharply. This phenomenon is likely due to the need to decode multiple masked tokens simultaneously at the start of a new block, which increases the model’s uncertainty.

\section{Limitations}
\label{app:sec_limitations}

\subsection{Discussions on Limitations}
\label{app:subsec_limitations}

While our temporal self-consistency voting and post-training approach demonstrates effectiveness in many scenarios, it exhibits significant limitations when applied to tasks where the model’s intermediate predictions are consistently inaccurate. 
As discussed in \cref{app:subsec_failure}, for the Sudoku dataset, the average correctness across all intermediate generation steps remains exceedingly low (below 5\%), making it difficult to reliably vote for the correct answer.
Similarly, RFT relies on the model already achieving reasonably good performance to produce meaningful reward signals~\citep{prabhudesai2025rent, agarwal2025unreasonable}, though combining TSE with the accuracy reward may alleviate this issue to some extent.
This underscores that our approach depends on the model’s inherent ability to generate correct or near-correct answers in the sampling trajectory.

\subsection{Failure Case Analysis}
\label{app:subsec_failure}

\begin{table}[t] 
\centering
\caption{
    \textbf{
        Performance of temporal self-consistency voting and temporal consistency reinforcement on the Sudoku dataset.
    }
    The baseline model is obtained by applying supervised fine-tuning to LLaDA-8B-Instruct using the s1K dataset.
}
\label{tab:results_sudoku}
\resizebox{0.5\textwidth}{!}{
\begin{tabular}{ll|ccc}
    \toprule
    \textbf{Model}
    & 
    & \multicolumn{3}{c}{\textbf{Sudoku}} \\
    \cmidrule(lr){3-5} 
    & \textbf{Method / Seq Len} & 128 & 256 & 512 \\
    \midrule
    & baseline & 12.2 & 6.7 & 5.5 \\
    \cmidrule(lr){2-5}
    & + temporal voting & 12.5 & 6.1 & 2.8 \\
    \midrule
    \multirow{4}{*}{+ RFT}
    & accuracy reward (d1) & 23.2 & 17.8 & 12.7 \\
    \cmidrule(lr){2-5}
    & negative TSE reward (ours) & 15.2 & 9.4 & 3.3 \\
    \cmidrule(lr){2-5}
    & combining both (ours) & \textbf{27.5} & \textbf{27.8} & \textbf{16.6} \\
    \bottomrule
\end{tabular}
}
\end{table}

\begin{figure}
    \centering
    \includegraphics[width=0.7\linewidth]{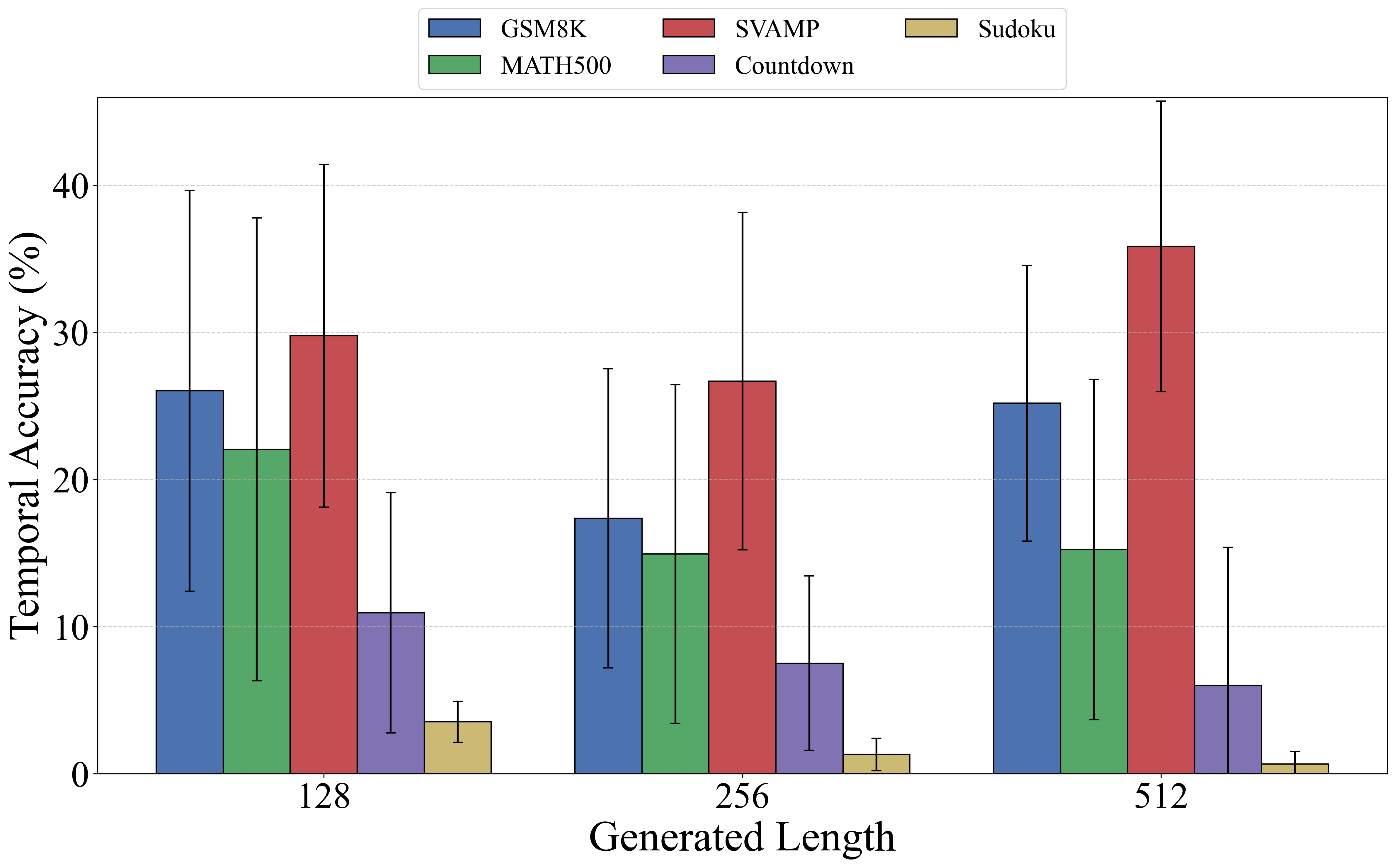}
    \caption{
        \textbf{Temporal accuracy across datasets.}
        Temporal Accuracy on Sudoku remains consistently low (all below 5\%) across different settings, indicating that the model rarely generates correct answers during the sampling process.
        This scarcity of valid candidates severely limits the effectiveness of voting mechanisms, as there are insufficient correct outputs to reliably converge on the correct answer.
    }
    \label{fig:app_temporal_accuracy}
\end{figure}

As discussed in \cref{app:subsec_limitations}, our method may rely on the model's initial performance to achieve further improvements. To investigate this limitation, we take the challenging Sudoku dataset as a case study.
As shown in \cref{tab:results_sudoku}, directly applying our proposed temporal self-consistency voting and temporal consistency reinforcement with the negative TSE reward alone results in a noticeable performance drop on the Sudoku dataset.
For example, the original accuracy of the base model for the generation length of 512 is 5.5\%, while the voting method achieves 2.8\%, reflecting a decline of 2.7\%.

To better understand this problem, we conduct a deeper analysis of the model's behavior during generation. We define a metric called \textbf{Temporal Accuracy}, which quantifies the average correctness across all intermediate sampling steps in the generation process. 
Formally, let $e_{i, t}$ represent the correctness indicator for the $i$-th question at the $t$-th sampling step, where $e_{i, t} = 1$ if the answer is correct, and $0$ otherwise.
Then, the \textbf{Temporal Accuracy} for a dataset with $N$ examples and $T$ sampling steps is computed as:
\begin{equation}
    \operatorname{TemporalAccuracy} = \frac{1}{N \cdot T} \sum_{i=1}^{N} \sum_{t=1}^{T}e_{i,t}.
\end{equation}
As shown in \cref{fig:app_temporal_accuracy}, the \textbf{Temporal Accuracy} of the Sudoku is much lower than the other 4 datasets, where temporal self-consistency voting proved effective.
On the Sudoku dataset, we observe that the Temporal Accuracy remains exceedingly low—below 5\% on average—across all intermediate steps. This low signal makes it difficult for temporal voting mechanisms to reliably identify the correct answer and for reinforcement learning signals to guide the model effectively.

Interestingly, while RFT with only negative TSE reward leads to poorer results, combining TSE with accuracy reward can achieve better performance than using accuracy reward alone. We hypothesize that this is because the integration of TSE allows the model to receive more fine-grained rewards, rather than just binary outcomes of correct or incorrect.

\section{More Experimental Results}
\label{app:more_exp}

\subsection{Training A Unified Model Across Multiple Tasks}
\label{app:subsec_joint_train}

To evaluate the generalization ability of our approach, we validate the proposed method in a multi-task setting. Specifically, we jointly train a single model on four datasets: GSM8K, MATH500, SVAMP, and Countdown. To ensure balanced training across tasks, we subsample each dataset so that the number of training examples is equal for all tasks.

\cref{table:combine_rl} summarizes the results. 
Across all four benchmarks, our combined reward method consistently outperforms the supervised fine-tuning baselines. 
On GSM8K, we observe clear gains at shorter sequence lengths, for example, with LLaDA-1.5 on length 128, accuracy improves from 69.8 to 73.7, yielding +3.9 points. 
On SVAMP, similar short-sequence gains are observed, 
% with LLaDA-1.5 at length 128 rising 
from 81.7 to 88.7, a +7.0 points increase. 
For the more challenging MATH500, the best improvement occurs at LLaDA-1.5, length 512, where performance increases from 35.4 to 41.6 (+6.2 points). 
Finally, on Countdown, our method achieves dramatic improvements, for instance, with LLaDA-1.5 at length 512, accuracy climbs from 20.7 to 50.4, a striking +29.7 points.

These results demonstrate that our method not only scales well across different reasoning domains but also enhances robustness under varying sequence lengths, confirming its effectiveness as a unified reward design for multi-task learning.

\begin{table}[t] 
\centering
\caption{
\textbf{Unified Model Performance Across Multiple Tasks.}
A single model is trained jointly on GSM8K, MATH500, SVAMP, and Countdown.
} 
\label{table:combine_rl} 
\vspace{2pt}
\resizebox{\textwidth}{!}{
\begin{tabular}{ll|cccccccccccc}
    \toprule
    &
    & \multicolumn{3}{c}{\textbf{GSM8K}} 
    & \multicolumn{3}{c}{\textbf{MATH500}} 
    & \multicolumn{3}{c}{\textbf{SVAMP}} 
    & \multicolumn{3}{c}{\textbf{Countdown}} \\
    \cmidrule(lr){3-5} \cmidrule(lr){6-8} \cmidrule(lr){9-11} \cmidrule(lr){12-14}
     & \textbf{Method / Seq Len} & 128 & 256 & 512 & 128 & 256 & 512 & 128 & 256 & 512 & 128 & 256 & 512 \\
    \midrule
    \multirow{1}{*}{\textbf{LLaDA}} 
    & baseline   & 70.2 & 78.7 & 80.1 & 23.8 & 34.4 & 36.8 & 81.7 & 83.3 & 82.7 & 21.5 & 19.9 & 21.5 \\
    \midrule
    \multirow{4}{*}{+ RFT}
    & accuracy reward (d1)  & 71.3 & 77.3 & 80.7 & 28.0 & 33.6 & 39.2 & 84.0 & 84.0 & 85.0 & 25.8 & 22.3 & 37.1 \\
    \cmidrule(lr){2-14}
    & \multirow{1}{*}{negative TSE reward (ours)} & 71.5 & 77.2 & \textbf{81.2} & \textbf{29.0} & 33.8 & 39.0 & 86.3 & 84.7 & \textbf{87.0} & 36.7 & 25.4 & 46.1 \\
    \cmidrule(lr){2-14}
    % \multirow{1}{*}{+ RFT}
    & \multirow{1}{*}{combined reward (ours)} & \textbf{72.4} & \textbf{78.8} & {80.4} & {28.8} & \textbf{35.2} & \textbf{40.2} & \textbf{88.7} & \textbf{87.0} & {86.0} & \textbf{37.1} & \textbf{27.7} & \textbf{46.9}  \\
    \midrule
    \midrule
    \multirow{1}{*}{\textbf{LLaDA-1.5}} 
    & baseline   & 69.8 & 79.4 & 81.1 & \textbf{29.0} & 32.4 & 35.4 & 85.3 & 86.3 & 83.3 & 21.5 & 21.1 & 20.7 \\
    \midrule
    \multirow{4}{*}{+ RFT}
    & accuracy reward (d1) & 72.5 & 79.2 & 81.7 & 28.6 & \textbf{35.0} & 39.0 & 89.0 & 87.7 & 85.3 & 30.5 & 24.6 & 41.8 \\
    \cmidrule(lr){2-14}
    & \multirow{1}{*}{negative TSE reward (ours)} & 72.5 & 79.2 & 81.4 & \textbf{29.0} & 32.6 & 38.2 & \textbf{89.3} & 85.3 & 86.7 & {37.3} & \textbf{44.5} & 34.0 \\
    \cmidrule(lr){2-14}
    % \multirow{1}{*}{+ RFT} 
    & \multirow{1}{*}{combined reward (ours)}
    & \textbf{73.7} & \textbf{79.5} & \textbf{82.5} & 28.2 & \textbf{35.0} & \textbf{41.6} & {88.3} & \textbf{90.0} & \textbf{88.0} & \textbf{37.9} & {29.7} & \textbf{50.4} \\
    \bottomrule
\end{tabular}
}
\end{table}

\subsection{Ablations on Scoring Rules for Combining TSE with Accuracy Reward}
\label{app:subsec_scoring_rules}

In \cref{subsec:method_rl}, we combine TSE with the accuracy reward using the proper scoring rule~\citep{gneiting2007strictly}. 
The purpose of these scoring rules is to promote truthful confidence assessments: they attain their minimum value when the predicted confidence $c$ precisely mirrors the actual probability that the model’s output $o$ corresponds to the correct answer $o^*$.
An exception is the spherical version we use, which consistently encourages both correctness and higher certainty, without penalizing overconfidence in incorrect predictions.
Here, we conduct an ablation study on different scoring rules. Specifically, we consider the following four forms:
\begin{equation*}
\begin{aligned}
    \textbf{Entropy scoring:} \quad & r^{\text{ent}}_{i} = \mathbbm{1}_{o_i = o^*} \cdot c(o_i) \\
    \textbf{Quadratic scoring:} \quad & r^{\text{quad}}_{i}= \mathbbm{1}_{o_i = o^*} -(c(o_i) - \mathbbm{1}_{o_i = o^*})^2 
    \\
    \textbf{Logistic scoring:} \quad & r^{\text{log}}_{i} = \mathbbm{1}_{o_i = o^*} + \mathbbm{1}_{o_i = o^*} \log (c(o_i)) + (1 - \mathbbm{1}_{o_i = o^*}) \log (1 - c(o_i))
    \\
    \textbf{Spherical scoring:} \quad & r^{\text{sphe}}_{i} = \mathbbm{1}_{o_i = o^*} + \frac{c(o_i)}{\sqrt{(c(o_i))^2  + (1 - c(o_i))^2}}
\end{aligned}
\end{equation*}
All four proposed reward functions aim to jointly encourage correctness and temporal self-consistency. Notably, the $r^{\text{ent}}_{i}$ function gives a reward of 0 for incorrect answers, while for correct answers, the reward is given by $c(o_i) = \frac{\mathcal{H}_{\max} - \operatorname{TSE}(o_i)}{\mathcal{H}_{\max}}$, where $\mathcal{H}_{\max} = \log T$. This reward reaches a maximum value of 1 when all sampling steps yield the correct answer.
The remaining three functions, $r^{\text{quad}}_{i}$, $r^{\text{log}}_{i}$, and $r^{\text{sphe}}_{i}$, correspond to the commonly used quadratic scoring, logarithmic scoring, and spherical scoring in proper scoring rules~\citep{gneiting2007strictly}, respectively.
We report the performance of all four reward combination methods in \cref{tab:results_rl_withgt}. By default, we use the spherical scoring rule because it demonstrates more superior results compared to the alternatives.

\begin{table*}[t]
    \centering
    \caption{
        \textbf{Ablations on scoring rules for combining TSE with accuracy reward.} We compare 4 different scoring rules, including entropy scoring, quadratic scoring, spherical scoring, and logistic scoring.
    }
    \scriptsize
    \setlength{\tabcolsep}{4pt}
    \resizebox{\textwidth}{!}{
    \begin{tabular}{ll|cccccccccccc}
        \toprule
        \textbf{Model / Dataset}
        &
        & \multicolumn{3}{c}{\textbf{GSM8K}} 
        & \multicolumn{3}{c}{\textbf{MATH500}} 
        & \multicolumn{3}{c}{\textbf{SVAMP}} 
        & \multicolumn{3}{c}{\textbf{Countdown}} \\
        \cmidrule(lr){3-5} \cmidrule(lr){6-8} \cmidrule(lr){9-11} \cmidrule(lr){12-14}
         & \textbf{Method / Seq Len} & 128 & 256 & 512 & 128 & 256 & 512 & 128 & 256 & 512 & 128 & 256 & 512 \\
        \midrule
        \multirow{1}{*}{\textbf{LLaDA}}
        & baseline   & 70.2 & 78.7 & 80.1 & 23.8 & 34.4 & 36.8 & 81.7 & 83.3 & 82.7 & 21.5 & 19.9 & 21.5 \\
        \midrule
        \multirow{4}{*}{+RFT}
        & entropy & 71.7 & 78.5 & 82.3 & 31.6 & \textbf{38.2} & 39.2 & \textbf{88.7} & 89.3 & 89.3 & {47.6} & \textbf{50.0} & {53.1} \\
        & quadratic & 71.3 & 79.9 & 82.0 & 31.0 & 37.6 & 40.0 & 88.0 & 8.3 & 89.3 & \textbf{50.0} & 34.4 & 48.1 \\
        & logistic & \textbf{73.0} & 79.5 & 81.1 & \textbf{31.6} & 36.8 & 39.4 & 87.7 & 87.3 & 90.7 & 46.5 & 37.9 & 51.2 \\
        & spherical & 72.1 & \textbf{80.0} & \textbf{83.0} & 31.2 & 35.4 & \textbf{41.4} & 85.0 & \textbf{90.3} & \textbf{92.3} & 41.5 & 42.6 & \textbf{54.7}  \\
        \midrule
        \midrule
        \multirow{1}{*}{\textbf{LLaDA-1.5}}
        & baseline   & 69.8 & 79.4 & 81.1 & 29.0 & 32.4 & 35.4 & 85.3 & 86.3 & 83.3 & 21.5 & 21.1 & 20.7 \\
        \midrule
        \multirow{4}{*}{+RFT}
        & entropy & 71.2 & 79.2 & 83.4 & 27.8 & \textbf{36.6} & 41.2 & 86.7 & \textbf{90.3} & 89.0 & 44.1 & 45.3 & 58.6 \\
        & quadratic & \textbf{73.4} & 79.1 & 83.2 & 29.2 & 33.2 & \textbf{42.4} & 86.0 & 86.7 & 90.0 & 43.4 & 45.3 & 57.4 \\
        & logistic & 71.2 & 79.2 & 83.1 & \textbf{30.2} & 33.8 & 41.0 & \textbf{88.3} & 89.0 & \textbf{91.0} & \textbf{48.8} & \textbf{46.9} & 61.3 \\
        & spherical & 73.2 & \textbf{80.5} & \textbf{84.0} & 29.6 & 35.4 & 41.4 & 86.3 & \textbf{90.3} & 89.0 & {44.5} & \textbf{46.9} & \textbf{63.3} \\
        \bottomrule
    \end{tabular}
    }
    \label{tab:results_rl_withgt}
\end{table*}

\subsection{\methodnamevoting After RFT}
\label{app:subsec_vote_after_rl}

\begin{table}[t] 
\centering
\caption{
\textbf{
Performance of temporal majority voting after reinforcement learning. 
}
Temporal self-consistency voting was applied to the model fine-tuned via temporal consistency reinforcement.
The upper part is derived from the LLaDA-8B-Instruct model trained using the Negative TSE reward, whereas the lower part is based on the model trained with a combination of TSE and accuracy reward.
For reference, we include the oracle $\operatorname{EverPass} @ 1 \mid t$ as a performance upper bound.
}
\label{tab:voting_after_rl} 
\resizebox{\textwidth}{!}{
\begin{tabular}{ll|cccccccccccc}
    \toprule
    &
    & \multicolumn{3}{c}{\textbf{GSM8K}} 
    & \multicolumn{3}{c}{\textbf{MATH500}} 
    & \multicolumn{3}{c}{\textbf{SVAMP}} 
    & \multicolumn{3}{c}{\textbf{Countdown}} \\
    \cmidrule(lr){3-5} \cmidrule(lr){6-8} \cmidrule(lr){9-11} \cmidrule(lr){12-14}
     & \textbf{Method / Seq Len} & 128 & 256 & 512 & 128 & 256 & 512 & 128 & 256 & 512 & 128 & 256 & 512 \\
    \midrule
    \multirow{1}{*}{\textbf{Negative TSE reward}} 
    & After RFT  & {72.2} & {78.8} & 80.2 & 30.6 & 34.6 & 38.0 & 84.3 & {89.0} & {88.7} & {38.6} & {53.5} & {44.9}  \\
    \midrule
    \multirow{4}{*}{+ Temporal Voting} & Fixed Weighting & 70.4 & 77.6 & 80.2 & 30.8 & 34.4 & 37.6 & 85.0 & 88.3 & 88.7 & 39.5 & 53.5 & 45.7 \\
    \cmidrule(lr){2-14}   
    & Linear Weighting & 72.3 & 78.8 & 80.6 & 30.8 & 35.4 & 38.0 & 85.3 & 88.3 & 88.7 & 39.1 & 53.5 & 45.7 \\
    \cmidrule(lr){2-14}
    & \multirow{2}{*}{Exp. Weighting} & 72.6 & 79.2 & 80.6 & 30.8 & 35.0 & 38.0 & 85.3 & 89.0 & 88.7 & 39.5 & 53.9 & 45.7 \\
    &  & \textcolor{green!70!black}{+0.4} & \textcolor{green!70!black}{+0.4} & \textcolor{green!70!black}{+0.4} & \textcolor{green!70!black}{+0.2} & \textcolor{green!70!black}{+0.4} & \textcolor{gray}{+0.0} & \textcolor{green!70!black}{+1.0} & \textcolor{gray}{+0.0} & \textcolor{gray}{+0.0} & \textcolor{green!70!black}{+0.9} & \textcolor{green!70!black}{+0.4} & \textcolor{green!70!black}{+0.8} \\
    \midrule
    & \demphs{$\operatorname{EverPass} @ 1 \mid t$}  & \demphs{80.5} & \demphs{81.8} & \demphs{81.3} & \demphs{33.4} & \demphs{37.4} & \demphs{40.0} & \demphs{88.7} & \demphs{90.7} & \demphs{89.7} & \demphs{44.1} & \demphs{59.8} & \demphs{55.5} \\
    \midrule
    \midrule
    \multirow{1}{*}{\textbf{TSE \& accuracy reward}} 
    & After RFT  & 72.1 & {80.0} & {83.0} & {31.2} & 35.4 & {41.4} & 85.0 & {90.3} & {92.3} & {41.5} & 42.6 & {54.7}  \\
    \midrule
    \multirow{4}{*}{+ Temporal Voting} & Fixed Weighting & 71.2 & 79.6 & 82.8 & 31.2 & 35.6 & 41.2 & 85.3 & 90.7 & 92.3 & 40.6 & 41.5 & 53.2 \\
    \cmidrule(lr){2-14}   
    & Linear Weighting & 73.0 & 81.9 & 83.0 & 31.2 & 36.0 & 41.0 & 86.7 & 90.3 & 92.7 & 40.9 & 42.8 & 54.2 \\
    \cmidrule(lr){2-14}
    & \multirow{2}{*}{Exp. Weighting} & 72.6 & 81.1 & 83.3 & 31.6 & 35.8 & 41.4 & 86.3 & 90.7 & 92.3 & 41.5 & 42.7 & 55.1 \\
    &  & \textcolor{green!70!black}{+0.5} & \textcolor{green!70!black}{+1.1} & \textcolor{green!70!black}{+0.3} & \textcolor{green!70!black}{+0.4} & \textcolor{green!70!black}{+0.4} & \textcolor{gray}{+0.0} & \textcolor{green!70!black}{+1.3} & \textcolor{green!70!black}{+0.4} & \textcolor{gray}{+0.0} & \textcolor{gray}{+0.0} & \textcolor{green!70!black}{+0.1} & \textcolor{green!70!black}{+0.4}  \\
    \midrule
    & \demphs{$\operatorname{EverPass} @ 1 \mid t$}  & \demphs{84.0} & \demphs{91.4} & \demphs{87.9} & \demphs{43.6} & \demphs{49.2} & \demphs{52.0} & \demphs{90.7} & \demphs{91.7} & \demphs{92.7} & \demphs{54.3} & \demphs{68.0} & \demphs{70.3} \\
    \bottomrule
\end{tabular}
}
\end{table}

It is worthwhile to investigate whether Temporal Self-consistency Voting continues to provide performance benefits after Temporal Consistency Reinforcement. 
To this end, we conducted experiments using two models: the first is derived from the LLaDA-8B-Instruct model trained with the negative TSE reward, and the second is trained using the accuracy reward combined with TSE.

We applied \methodnamevoting to both models, and the results are summarized in \cref{tab:voting_after_rl}. The findings indicate that both models still exhibit performance improvements when temporal voting is applied, even after undergoing reinforcement learning. 
This suggests that \methodnamevoting and \methodnamerft are complementary techniques, and can be effectively combined to further enhance model performance.

\subsection{Examples of time oscillation}
\label{app:subsec_examples_oscillation}

We present representative examples of temporal oscillation from GSM8K, using LLaDA-8B-Instruct as the backbone model. 
Note that all examples shown below were ultimately incorrect but were classified as correct answers by our voting method, as described in \cref{subsec:method_voting}.
Correct answers are in \textcolor{green!70!black}{green}, incorrect answers in \textcolor{red}{red}.
We use \textcolor{blue}{blue} to highlight the key segments in each example.

We observe that many questions become incorrect during temporal oscillation due to flawed reasoning paths. Although the model may initially provide the correct answer and follow a logical reasoning path, an incorrect trajectory can ultimately lead to an erroneous response.

A notable case is seen in Example 3. Initially, the model correctly calculates the total cups, per-hour revenue, and the per-hour cost deduction to derive the right profit per hour (step 54). However, during reasoning, it introduces erroneous masking regarding key calculation elements—the per-hour cost to subtract. Even though the answer remains correct at first (step 61), the flawed reasoning trajectory eventually results in miscalculations and an incorrect final response (step 62).

\section{Change Log}
\label{app:change_log}

Here we list the main modifications introduced in \textit{arXiv v2} compared to \textit{v1}:
\begin{itemize}
    \item \textbf{Bug Fix in LLaDA-1.5 Evaluation.} 
    We identified a bug in the previous experiments on {LLaDA-1.5}, where LoRA modules trained on {LLaDA-1.5} were mistakenly evaluated by loading them into the {LLaDA} model. 
    Interestingly, this setting revealed a certain degree of generalization, as the LoRA trained on {LLaDA-1.5} still improved {LLaDA}'s performance. 
    In the current version, we corrected this bug by properly evaluating LoRA with the {LLaDA-1.5} backbone. 
    As a result, both our method and the reproduced baseline {d1} show improved performance (see \cref{table:main_rl} and \cref{tab:results_rl_withgt}).

    \item \textbf{Unified Multi-Task Training.} 
    We added new results for \emph{Training a Unified Model Across Multiple Tasks}, as presented in~\cref{app:subsec_joint_train}.

    \item \textbf{Text Revision.} 
    The manuscript was revised to improve readability, with the main content condensed to nine pages for clarity and precision.
\end{itemize}

Here we summarize the main updates introduced in \textit{arXiv v3} compared to \textit{v2}:
\begin{itemize}
    \item \textbf{New Sections.} 
    Added dedicated sections on \textit{LLM Usage} and the \textit{Ethics Statement}.

    \item \textbf{Revised Text.} 
    The manuscript has been refined for improved readability and expanded with a more comprehensive discussion of related work.
\end{itemize}

\clearpage

\begin{tcolorbox}[colback=gray!5, colframe=gray!80!black, title=Example 1]

\textbf{Problem:}
Ruby is 6 times older than Sam. In 9 years, Ruby will be 3 times as old as Sam. How old is Sam now?

\medskip

\includegraphics[width=\linewidth]{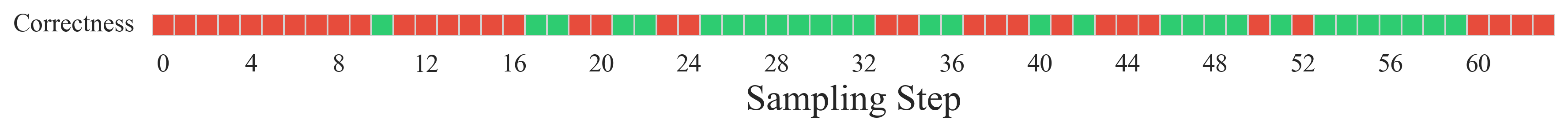}

\medskip

\textbf{Step 59:}

$<$reasoning$>$

Let Sam's age be $x$ years. Then Ruby's age is $6x$ years. In 9 years, Ruby's age will be $6x + 9$ and Sam's age will be $x + 9$.
According to the problem, in 9 years, Ruby will be 3 times as old as Sam, so we have the equation $6x + 9 = 3(x + 9)$. 
\textcolor{blue}{Simplifying this equation, we get $3x = 0$ so $x = 3$.}

$<$/reasoning$>$

$<$answer$>$

\textcolor{green!70!black}{\boxed{6}}

$<$/answer$>$

\medskip

\textbf{Step 60:}

$<$reasoning$>$

Let Sam's age be $x$ years. Then Ruby's age is $6x$ years. In 9 years, Ruby's age will be $6x + 9$ and Sam's age will be $x + 9$.
According to the problem, in 9 years, Ruby will be 3 times as old as Sam, so we have the equation $6x + 9 = 3(x + 9)$. 
\textcolor{blue}{Simplifying this equation, we get $3x = 0 \Rightarrow x = 0$}

$<$/reasoning$>$

$<$answer$>$

\textcolor{red}{\boxed{0}}

$<$/answer$>$

\end{tcolorbox}

% Example 2
\begin{tcolorbox}[colback=gray!5, colframe=gray!80!black, title=Example 2]

\textbf{Problem:}
Shiela bought five cell phones for \$150 each for a 3-month installment. A 2\% interest will be charged for each unit. How much will Shiela pay each month for 3 months

\medskip

\includegraphics[width=\linewidth]{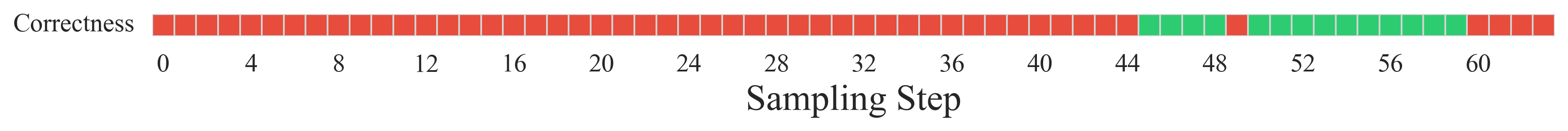}

\medskip

\textbf{Step 59:}

$<$reasoning$>$

Shiela bought five cell phones for \$150 each, so the total cost is $\$5 \times 150 = \$750$.
The interest for each unit is 2\%, so the total interest is $\$750 \times 0.02 = \$15$. 
SinceIELa will pay this amount over 3 months, the monthly payment is \$750 + \$15 = \$$<<.>>750 \backslash 1. =255.$

$<$/reasoning$>$

$<$answer$>$

\textcolor{green!70!black}{\boxed{255}}

$<$/answer$>$

\medskip

\textbf{Step 60:}

$<$reasoning$>$

Shiela bought five cell phones for \$150 each, so the total cost is $\$5 \times 150 = \$750$.
The interest for each unit is 2\%, so the total interest is $\$750 \times 0.02 = \$15$. 
SinceIELa will pay this amount over 3 months, the monthly payment is \$750 + \$15 = \$\textcolor{blue}{$<<.>>750/3 =>>250$}.

$<$/reasoning$>$

$<$answer$>$

\textcolor{red}{\boxed{250}}

$<$/answer$>$

\end{tcolorbox}

% 示例 3
\begin{tcolorbox}[colback=gray!5, colframe=gray!80!black, title=Example 3]
\textbf{Problem:} 
Millie decides to open a lemonade stand. She spends \$18 to buy enough supplies to make 3 pitchers of lemonade. Each pitcher holds 12 cups of lemonade. She sells each cup of lemonade for \$1. She sells an average of 4 cups per hour that her lemonade stand is open. If Millie sells all of the lemonade, how much profit will she make per hour that she spends running the lemonade stand?

\medskip

\includegraphics[width=\linewidth]{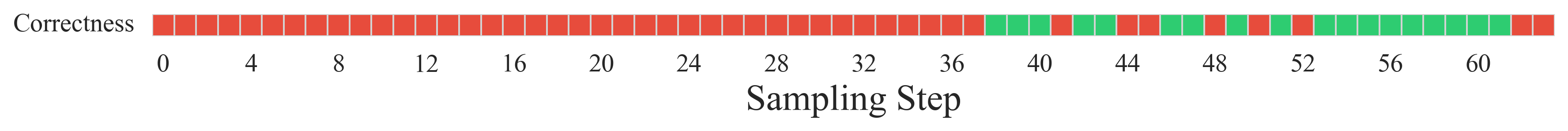}

\medskip
\textbf{Step 54:}

$<$reasoning$>$

Millie spends \$18 on supplies to make 3 pitchers of lemonade. Each pitcher holds 12 cups, so she has a total of 3 * 12 = 36 cups of lemonade.
She sells each cup for \$1 and and sells an average of 4 cups per hour. Therefore, her revenue per hour is 4 * \$1 = \$4.
\textcolor{blue}{To find per hour profit, we subtract the cost of supplies from the revenue: \$4 - \$2 = \$2 per hour.}

$<$/reasoning$>$

$<$answer$>$

\textcolor{green!70!black}{\boxed{2}}

$<$/answer$>$

\medskip

$\cdots$

\medskip

\textbf{Step 61:} 

$<$reasoning$>$

Millie spends \$18 on supplies to make 3 pitchers of lemonade. 
Each pitcher holds 12 cups, so she has a total of 3 * 12 = 36 cups of lemonade. 
She sells each cup for \$1 and and sells an average of 4 cups per hour. 
Therefore, her revenue per hour is 4 * \$1 = \$4. 
\textcolor{blue}{To find per hour profit, we subtract the cost of supplies from the revenue: \$4 - \$3 = \$2 per hour.}

$<$/reasoning$>$

$<$answer$>$

\textcolor{green!70!black}{\boxed{2}}

$<$/answer$>$

\medskip

\textbf{Step 62:} 

$<$reasoning$>$

Millie spends \$18 on supplies to make 3 pitchers of lemonade. 
Each pitcher holds 12 cups, so she has a total of 3 * 12 = 36 cups of lemonade. 
She sells each cup for \$1 and and sells an average of 4 cups per hour. 
Therefore, her revenue per hour is 4 * \$1 = \$4. 
\textcolor{blue}{To find per hour profit, we subtract the cost of supplies from the revenue: \$4 - \$3 = \$1 per hour.}

$<$/reasoning$>$

$<$answer$>$

\textcolor{red}{\boxed{1}}

$<$/answer$>$

\end{tcolorbox}

\end{document}